\documentclass[12pt,a4paper]{article}

\usepackage{graphicx}

\usepackage{amsmath}
\usepackage{graphicx}
\usepackage{amssymb}
\usepackage{color}
\usepackage{enumerate}

\usepackage{geometry}
\newenvironment{definition}[1] []{\noindent{\emph{#1.}}}

\newenvironment{acknowledgements}[1][]{\vspace{0.5cm} \noindent \hspace{-0.4cm} \textbf{\fontsize{5.6mm}{7mm} \selectfont Acknowledgements.}~#1}

\definecolor{antiquewhite}{rgb}{0.98, 0.92, 0.84}

\newcommand{\highlight}[1]{\colorbox{antiquewhite}{\parbox{0.9\textwidth}{#1}}}

\newcommand{\seenote}{\footnotemark \hspace{0.1cm}} % see note ... 
\newcommand{\note}[1]{\footnotetext{#1}} % text of the note

\newcommand{\down}{\vspace{0.2cm}}
\newcommand{\Down}{\vspace{0.4cm}}
\newcommand{\DDown}{\vspace{1cm}}

\newcommand{\up}{\vspace{-0.2cm}}
\newcommand{\Up}{\vspace{-0.4cm}}

\newcommand{\emh}[1]{\mathcal{#1}}
\newcommand{\lt}{<}
\newcommand{\gt}{>}
\newcommand{\mathdef}{\stackrel{\triangle}{=}} 
\newcommand{\dsum}[3]{\displaystyle \sum_{#1}^{#2}{#3}}
\newcommand{\dprod}[3]{\displaystyle \prod_{#1}^{#2}{#3}}
\newcommand{\infinity}{\infty}

\newcommand{\instanceof}{\triangleleft}

\newcommand{\prm}{^\prime}
\newcommand{\what}[1]{\widehat{#1}}
\newcommand{\ALIGN}[1]{\begin{align} #1 \end{align}}
\newcommand{\ALIGNX}[1]{\begin{align*} #1 \end{align*} }
\newcommand{\tok}[2][k]{#2^{(#1)}} 
\newcommand{\bmx}[1]{\begin{bmatrix} #1 \end{bmatrix}}
\newcommand{\cmx}[1]{\begin{bmatrix} #1_{00} & #1_{01} \\ #1_{10} & #1_{11} \end{bmatrix}}
\newcommand{\diagcmx}[2]{\begin{bmatrix} #1 & 0 \\ 0 & #2 \end{bmatrix}}
\newcommand{\zerocmx}{\begin{bmatrix} 0 & 0 \\ 0 & 0 \end{bmatrix}}
\newcommand{\identitycmx}{\begin{bmatrix} 1 & 0 \\ 0 & 1 \end{bmatrix}}
\newcommand{\neutralcmx}{\begin{bmatrix} 0 & 1 \\ 0 & 1 \end{bmatrix}}
\newcommand{\innercmx}[2]{\begin{bmatrix} 0 & #1 \\ 0 & #2 \end{bmatrix}}
\newcommand{\outercmx}[2]{\begin{bmatrix} #1 & 0 \\ #2 & 0 \end{bmatrix}}
\newcommand{\Oplus}[2]{\begin{bmatrix} #1_{00} && 0 \\ #1_{10} && 0 \end{bmatrix} + 
                                         \begin{bmatrix} #1_{01} && 0 \\ 0 && #1_{11} \end{bmatrix} \cdot
                                         \begin{bmatrix} #2_{00} && #2_{01} \\ #2_{10} && #2_{11} \end{bmatrix}}
\newcommand{\OOOplus}[2]
   {\begin{bmatrix} #1_{00} + #1_{01} \cdot #2_{00} && #1_{01} \cdot #2_{01} \\
                              #1_{10} + #1_{11} \cdot #2_{10} && #1_{11} \cdot #2_{11} \end{bmatrix} }

\newcommand{\milestone}[1]{}
\newcommand{\comment}[1]{}

\newcommand{\submilestone}[1]{}

\newcommand{\ITEMIZE}[1]{\begin{itemize} #1 \end{itemize}}

\newcommand{\HC}{\emph{HC}}
\newcommand{\HTC}{\emph{HTC}}

\newcommand{\binrel}{\le}
\newcommand{\taxonomy}[1][T]{\emh{#1} = \langle \emh{C}, \binrel \rangle}
\newcommand{\kleenestar}[1]{\emh{#1}^{*}}
\newcommand{\kleenestarP}[1]{\emh{#1}^{+}}
\newcommand{\wfs}[1][w]{#1}
\newcommand{\pipe}{ \pi = c_0 c_1 \ldots c_L}

\newcommand{\totalex}{m}
\newcommand{\posex}{p}
\newcommand{\negex}{n}

\newcommand{\fk}[1][k]{f_{#1}}
\newcommand{\nfk}[1][k]{\bar{f}_{#1}}
\newcommand{\Fk}[1][k]{F_{#1}}
\newcommand{\nFk}[1][k]{\bar{F}_{#1}}
\newcommand{\Xk}[1][k]{X_{#1}}
\newcommand{\Ck}[1][k]{\widehat{X}_{#1}}
\newcommand{\ok}[1][k]{c_{#1}}
\newcommand{\ck}[1][k]{\widehat{c}_{#1}}
\newcommand{\ext}[2][k]{#2_{#1}}
\newcommand{\extw}[2][k]{\what{#2}_{#1}}
\newcommand{\neutral}{\mu}

\newcommand{\jointk}[1][k]{p(X_{#1}, \what{X}_{#1})}

\begin{document}
\setcounter{page}{1}
%\issue{???~(2013)}

\title{Probabilistic Modelling of Progressive Filtering}

%\address{Address for correspondence goes here}

\author{Giuliano Armano\\
Dept. of Electrical and Electronic Engineering\\
University of Caliari\\ Piazza d'Armi, 09123, Cagliari, Italy\\
armano{@}diee.unica.it}
%\and Second Author\thanks{Thanks for something else to somebody else}\\
%Department of Informatics \\
%City University \\
%London, England }

\date{}

\maketitle

%\runninghead{G. Armano}{Modelling PF}

\highlight{The article entitled \emph{Modeling Progressive Filtering}, published on Fundamenta Informaticae (Vol. 138, Issue 3, pp. 285-320, July 2015), has been derived from this extended report.}

\begin{abstract}

Progressive filtering is a simple way to perform hierarchical classification, inspired by the behavior that most humans put into practice while attempting to categorize an item according to an underlying taxonomy.
Each node of the taxonomy being associated with a different category, one may visualize the categorization process by looking at the item going downwards through all the nodes that accept it as belonging to the corresponding category.
This paper is aimed at modeling the progressive filtering technique from a probabilistic perspective, in a hierarchical text categorization setting.  
As a result, the designer of a system based on progressive filtering should be facilitated in the task of devising, training, and testing it.

%\keywords{Information Filtering \and Hierarchical Classification \and Hierarchical Text Categorization}

\end{abstract}

%\begin{keywords}
%Hierarchical Classification, Information Filtering, Text Categorization
%\end{keywords}

%\newcounter{def_counter}

%\newtheorem{definition} [def_counter]{Def.}

\section{Introduction} \label{sec:introduction} \milestone{INTRODUCTION}

Classification (or categorization) is a process of labeling data with categories taken from a predefined set, supposed to be semantically relevant to the problem at hand. The absence of an internal structure in the set of categories (or the absence of techniques able to account for this structure) leads to the so-called ``flat'' models, in which categories are dealt with independently of one another.
In the event that the categories are organized in a taxonomy, typically through \emph{is-a} or \emph{part-of} relationships, and assuming that one wants to take into account also this information in order to improve the performance of a categorization system, the corresponding labeling process takes the name of hierarchical classification ($\HC$). This research area has received much attention after the explosion of the World Wide Web, in which many problems and the corresponding software applications are based on an underlying taxonomy (e.g., web search with ``vertical'' search engines, online marketplaces, recommender systems).

This paper is aimed at modeling progressive filtering (hereinafter PF), a hierarchical technique inspired by the behavior that most humans put into practice while attempting to categorize data according to a taxonomy. PF assumes that a top-down categorization process occurs, performed in combination with a set of binary classifiers that mirror the structure of the taxonomy and are entrusted with accepting relevant inputs while rejecting the others. 
Starting from the root, supposed to be unique, a classifier that accepts an input passes it down to all its offspring (if any), and so on. The typical result consists of activating one or more paths within the taxonomy, i.e., those for which the corresponding classifiers have accepted the given input. 
While concentrating on hierarchical text categorization ($\HTC$) problems, we will be focusing on the following issues: i) Can we predict the expected behavior of a system implemented in accordance with PF when fed with a corpus of documents whose statistical properties are known?, ii) Would it be feasible to separate the statistical information concerning inputs from the intrinsic properties of the classifiers embedded in the given taxonomy?

To my knowledge, no previous work has been done on the above issues, although the reasons for investigating them from a probabilistic perspective are manifold. In particular, a probabilistic model able to estimate the outcomes of a system that implements PF when applied to a real-world task can facilitate taxonomy design, optimization, and assessment.
As for \emph{taxonomy design}, the ability to assess in advance an update to the underlying taxonomy could be very useful for a designer. Indeed, adding or removing a node, as well as updating its characteristics, can have a substantial impact on performance, also depending on which metrics the designer wants to maximize. The importance of the model is also motivated by the fact that nowadays many e-businesses (e.g., online stores) resort to human experts to create and maintain the taxonomies considered relevant for their business activities, mainly due to the lack of automatic or semi-automatic tools for taxonomy handling.
As for \emph{taxonomy optimization}, let us assume that each classifier in the taxonomy has some parameters for controlling its behavior. The simplest scenario consists of focusing on the acceptance threshold, which typically falls in the range $[0,1 ]$. According to this hypothesis, an optimization problem over the threshold space arises, characterized by high time complexity.
To make this problem tractable, in our view, one should accept suboptimal solutions while looking at the above issues from a perspective based on three layers (listed from top to bottom): (i) the space of thresholds, (ii) the space of classifiers, and (iii) the space of experiments. 
For each layer, a source of intractability holds, which can be dealt with by means of approximate models. 
In particular, a ``light'' (hence, suboptimal) algorithm for threshold optimization can be used to search the space of thresholds; the mapping between the threshold and the expected behavior of a classifier can be estimated in the space of classifiers, \seenote
and the probabilistic model introduced in this paper can be used to predict the outcome of a test run on a corpus of documents with known statistical properties.
\note{Many subtle problems arise in the space of classifiers when trying to reflect a change imposed on the space of thresholds. As discussion of these issues is far beyond the scope of this paper, we limit our assertion to the generic recommendations above --intended to overcome the computational issues arising from the need to retrain classifiers.}
As for \emph{taxonomy assessment}, the possibility of evaluating relevant metrics can provide useful information about the expected behavior of a taxonomy. In particular, checking how the distribution of inputs, together with the characteristics of the embedded classifiers, affect the overall performance of a system compliant with PF can be very important while testing and maintaining a taxonomy.

The rest of the paper is organized as follows:
Section~\ref{sec:related-work} briefly recalls some related work, useful for fitting the problem within the current state-of-the-art.
Section~\ref{sec:htc} introduces the concepts deemed most relevant for $\HTC$.
Section~\ref{sec:progressive-filtering} defines PF, first from a probabilistic perspective and then as a linear transformation in the space of (normalized) confusion matrices.
Section~\ref{sec:metrics} analyzes how relevant metrics change within a taxonomy.
Section~\ref{sec:discussion} provides a critical assessment of PF.
Conclusions and future work (Section~\ref{sec:conclusions}) end the paper.

\section{Related Work} \label{sec:related-work} \milestone{RELATED WORK}

In line with the ``divide and conquer'' philosophy, the main advantage expected from the hierarchical perspective is that the problem is partitioned into smaller subproblems, hopefully easier than the original one, so that each can be effectively and efficiently managed.
Beyond this generic consideration, a number of algorithmic and architectural solutions have been experimented. A first rough division can be made between the so-called \emph{local}  vs. \emph{global} approach.
In the former case an ensemble of classifiers is generated, whereas in the latter a monolithic classifier is generated, able to account for the whole taxonomy. 
Local approaches seem to interpret the divide and conquer philosophy more properly, as they concentrate on (a typically small) part of the underlying taxonomy while implementing each component of the ensemble.
However, the global approach does not prevent local strategies from actually being used to generate a monolithic classifier (e.g., multi-label decision trees).

\subsection{Pachinko vs. Probabilistic Machines} \label{sub:pachinko} \milestone{PACHINKO}

In \cite{bib:kiritchenko06}, all local approaches that rely on a sequence of top-down decisions take the esoteric name of \emph{pachinko machine}, as they resemble to some extent the corresponding Japanese game. 
This approach has been widely used with different learning algorithms and techniques: linear classifiers \cite{bib:ng97}, \cite{bib:dalessio00}, probabilistic classifiers \cite{bib:koller97}, decision rules \cite{bib:ipeirotis01}, boosting \cite{bib:esuli08}, artificial neural networks (ANNs) \cite{bib:ruiz02}, support vector machines (SVMs) \cite{bib:sun01}, and in a transductive setting \cite{bib:ceci08}. Moreover, in \cite{bib:kiritchenko06}, an extended version of the Pachinko-machine approach is proposed, adding the ability to terminate the categorization process at any intermediate level of the hierarchy.

The so-called \emph{probabilistic machines} adopt an alternative approach, in which all paths are considered simultaneously. Their probabilities are calculated as the product of individual probabilities of categories (for each path), and the leaf categories (i.e., the most probable paths) are selected according to a maximum likelihood criterion. This approach has been used in combination with probabilistic classifiers, \cite{bib:chakrabarti97}, with ANNs \cite{bib:wiener95}, \cite{bib:weigend99}, and with SVMs \cite{bib:barutcuoglu06}.

It is worth pointing out that Dumais and Chen compared the two local approaches, i.e., Pachinko machine and probabilistic, and found no difference in performance, \cite{bib:dumais00}.

\subsection{Mapping Between Classifiers and the Underlying Taxonomy} \label{sub:mapping} \milestone{MAPPING}

According to the survey paper of %Silla and Freitas
\cite{bib:silla10}, a hierarchical approach is better understood when described from two dimensions, i.e., the nature of the given \emph{problem} (or class of problems) and the characteristics of the \emph{algorithm} devised to cope with it (or them).
The problem is described by a triple $\langle \Upsilon, \Psi, \Phi \rangle$, where:
$\Upsilon$ specifies the type of graph representing the hierarchical classes (i.e., tree or DAG),
$\Psi$ indicates whether a data instance is allowed to have class labels associated with a single or multiple paths in the taxonomy, and
$\Phi$ describes the label depth of the data instances, i.e., full or partial.
The algorithm is described by a 4-tuple $\langle \Delta, \Xi, \Omega, \Theta \rangle$, where:
$\Delta$ indicates whether single or multiple path prediction is performed,
$\Xi$ specifies whether leaf-node prediction is mandatory or not, 
$\Omega$ is the taxonomy structure the algorithm can handle (i.e., tree or DAG), and
$\Theta$ establishes the mapping between classifiers and the underlying taxonomy (i.e., local classifier per node, local classifier per parent node, local classifier per level, and global classifier).

A simple way to categorize the various proposals made in $\HC$
is to focus on the mapping between classifiers and the underlying taxonomy. 
Relevant proposals are listed from fine to coarse granularity:

\ITEMIZE{
\item \emph{Local Classifier per Node}.
This approach admits only binary decisions, as each classifier is entrusted with deciding whether the input at hand can be forwarded or not to its children.
\cite{bib:dalessio00}, \cite{bib:dumais00}, and \cite{bib:sun01} are the first proposals in which sequential Boolean decisions are applied in combination with local classifiers per node. In \cite{bib:wu05}, the idea of mirroring the taxonomy structure through binary classifiers is clearly highlighted (the authors call this technique ``binarized structured label learning'').
In \cite{bib:addis10KDIR}, the underlying taxonomy is scattered on the corresponding set of admissible paths which originate from the root (called pipelines).
Each component of a pipeline embeds a binary classifier, and pipelines are independently optimized.
\item \emph{Local Classifier per Parent Node}.
In the seminal work by %Koller and Sahami
\cite{bib:koller97}, a document to be classified proceeds top-down along the given taxonomy, each classifier being used to decide to which subtree(s) the document should be sent to, until one or more leaves of the taxonomy are reached. This approach, which requires the implementation of multiclass classifiers for each parent node, gave rise to a variety of actual systems, e.g., \cite{bib:ng97}, \cite{bib:dalessio00},\cite{bib:wibowo02a}, and \cite{bib:ruiz02}.
\item \emph{Local Classifier per Level}.
This approach can be considered as a boundary between local and global approaches, as the number of outputs per level grows moving down through the taxonomy, soon becoming comparable with the number required for a global classifier. Among the proposals adopting this approach, let us recall \cite{bib:kriegel04} and \cite{bib:clare03}.
\item  \emph{Global Classifier}. One classifier is trained, able to discriminate among all categories. Many global approaches to $\HC$ have been proposed, e.g.,  \cite{bib:wang99}, \cite{bib:wang01}, \cite{bib:itskevitch01}, \cite{bib:dekel04}, \cite{bib:tsochantaridis04}, \cite{bib:cai04}, and \cite{bib:kiritchenko06}.
}

According to \cite{bib:sun01}, training systems with a global approach is computationally heavy, as they typically do not exploit different sets of features at different hierarchical levels, and are not flexible, as a classifier must be retrained each time the hierarchical structure changes. On the other hand, although computationally more efficient, local approaches have to make several correct decisions in a row to correctly classify one example, and errors made at top levels are usually not recoverable. Moreover, the categories may lack positive examples at deep levels, making the task of training reliable classifiers difficult.

\subsection{Further Relevant Issues for HC} \label{sub:further-issues}
Further relevant issues for $\HC$ are the way feature selection/reduction is performed and which strategy is adopted to train the classifier(s) embedded in a hierarchical system. Research efforts in this area have focused largely on $\HTC$.

Features can be selected according to a global or a local approach (a comparison between the two approaches can be found in \cite{bib:weigend99}).
In global approaches, the same set of features is used at any level of the taxonomy, as done with flat categorization. 
This solution is normally adopted in monolithic systems, where only one classifier is entrusted with distinguishing among all categories in a taxonomy \cite{bib:frommholz01,bib:itskevitch01}. Variations on this theme can be found in \cite{bib:wibowo02a} and in \cite{bib:mccallum98}. 
In local approaches, different sets of features are selected for different nodes in the taxonomy, thus taking advantage of dividing a large initial problem into subproblems, e.g., \cite{bib:wibowo02a}. This is the default choice for Pachinko machines.
In a more recent work, \cite{bib:esuli08} suggest that feature selection should pay attention to the topology of the classification scheme. Among other approaches to feature selection, let us recall \cite{bib:liu95}, based on $\chi$-square feature evaluation. 
As for feature reduction, latent semantic indexing \cite{bib:deerwester90} is the most commonly used technique. 
Based on singular value decomposition \cite{bib:golub65}, it implements the principle that words used in the same contexts tend to have similar meanings.

As for training strategies, according to \cite{bib:ceci07}, training sets can be \emph{hierarchical} or \emph{proper}.
The former include documents of the subtree rooted in a category as positive examples and documents of the sibling subtrees as negative examples. 
The latter include documents of a category as positive examples (while disregarding documents from its offspring), and documents of the sibling categories as negative examples.
After running several experiments aimed at assessing the pros and cons of the two training strategies, the authors have shown that hierarchical training sets are more effective.

\section{Hierarchical Text Categorization} \label{sec:htc}  \milestone{HTC}

As our work will focus mainly on $\HTC$, let us summarize the basic concepts and the issues considered most relevant to this research field (see also \cite{bib:sebastiani02} and \cite{bib:kiritchenko06}).
\subsection{Standard Definitions for HTC}

\begin{definition}[Text Categorization] \label{def:text-categorization} 
Text categorization is the task of assigning a Boolean value to each pair $\langle d_{j}, c_{i}\rangle \in D \times C$, where $D$ is a domain of documents and $C= \{ \ok \; | \; k=1,2, ..., N\}$ is a set of \emph{N} predefined categories.
\end{definition} \down

\begin{definition}[Hierarchical Text Categorization] \label{def:htc}
Hierarchical Text Categorization is a text categorization task performed according to a given taxonomy $\taxonomy$, where $C= \{c_{k} \; | \; k=1,2, ..., N\}$ is a set of \emph{N} predefined categories and ``$\binrel$'' is a reflexive, anti-symmetric, and transitive binary relation.\seenote
\end{definition} \down
\note{~Some authors use ``$\lt$'' instead of ``$\binrel$'' as default binary relation. As the definition of ``$=$'' and ``$\lt$'' from ``$\binrel$'' is trivial, in the following we will use ``$\lt$'' when deemed useful for rendering definitions more intuitive.}

In the most general case, $\emh{T}$ can be thought of as a strict partially ordered set (strict poset), which can be graphically represented by a DAG. 
We assume known all ordinary definitions concerning posets. However, for the sake of readability, let us recall some relevant definitions. \down

\begin{definition} [Covering Relation] \label{def:covering-relation}
Given a taxonomy $\taxonomy$, the covering relation ``$\prec$'' holds between comparable elements that are immediate neighbors in the taxonomy.
In symbols:
$b \prec a \Leftrightarrow b \lt a \; \wedge \neg \; \exists \, c \in \emh{C} \; \text{s.t.} \; b \lt c \lt a$.
The characteristic function $f: \emh{C} \times \emh{C} \rightarrow [0,1]$ for the covering relation ``$\prec$'' is defined as:
\ALIGN{ \label{eq:char-f}
          f(b,a)=\left\{ \begin{array}{ll}
	1 & \text{if}\; \; b \prec a \\
	0 & \text{otherwise}\end{array}\right.
}
\end{definition} \down

A ``soft'' version of the above definition would substitute ``1'' (used to denote full membership) with a number intended to measure  to what extent the pair in question satisfies the covering relation. In a probabilistic setting, a natural choice for the characteristic function would be to let it coincide with the conditional probability $p(b \vert a)$. In symbols:
\ALIGN{
\forall a, b \in \emh{C}: \; b \prec a \iff f(b,a) \equiv p(b \vert a) > 0
}

\begin{definition}[Ancestors, Offspring, and Children Sets] \label{def:ancestors-etc}
The notions of ancestors, offspring, and children sets, useful when dealing with taxonomies, can be easily defined for posets (hence, for DAGs and trees).
Given a node $r \in \emh{C}$:

\Up

\ALIGN { 
\emh{A}(r) &= \{a \in C \mid r \lt a \}  	&& \text{$\emh{A}$ncestors set} \notag\\
\emh{O}(r) &= \{o \in C \mid o \lt r \} 		&& \text{$\emh{O}$ffspring set} \label{eq:ancestors-etc}\\
\emh{H}(r) &= \{c \in C \mid c \prec r \}     && \text{C$\emh{H}$ildren set} \notag
}
\end{definition} \down

\begin{definition} [Root, internal nodes, and leaves] \label{def:root-etc}
A category without ancestors is called \emph{root};
a category without children is called \emph{leaf}, and a category with both ancestors and offspring is called \emph{internal category}.
\end{definition} \down

\noindent Two constraints must be effective for hierarchical text categorization:
\ITEMIZE {
\item{\emph{Hierarchical Consistency.}}
A label set $C_{d} \subseteq \emh{C}$ assigned to an instance $d \in D$ is said to be consistent with a given taxonomy $\taxonomy$ if it includes the complete ancestor sets for every label $c \in C_{d}$.
In symbols: $c \in C_{d} \wedge b \in \emh{A}(c) \rightarrow b \in C_{d}$.
\down
\item{\emph{Hierarchical Consistency Requirement.}}
Any label assignments produced by a hierarchical classification system on a given categorization task has to be consistent with the underlying category taxonomy.
}

The notion of domain of a category $c$ is also relevant, which denotes all documents that belong to $c$ (i.e., the set of its positive instances).

\begin{definition} [Domain of a category] \label{def:domain}
Given a taxonomy $\taxonomy$ the domain of a category $c \in \emh{C}$ is defined as: \seenote
\ALIGN {  \label{eq:domain}
dom(\ok[])= \left\{ i \mid i \instanceof c \vee \exists a \in \emh{O}(c) \; s.t. \; i \instanceof a\right\}
}
\end{definition} \down

\note{Where ``$i \instanceof c$'' denotes the \emph{instance-of} relation that holds between the instance $i$ and the category $c$.}

We assume that each category $c \in \emh{C}$ embeds a corresponding binary classifier. 
Given an input, the classifier is entrusted with deciding whether or not it belongs to the corresponding category. 
To distinguish between a category and its embedded classifier, the latter will be denoted by a circumflex (i.e., $\ck[]$ denotes the classifier embedded by the category $\ok[]$).

\begin{figure}
\centering
\includegraphics[width=0.45\textwidth]{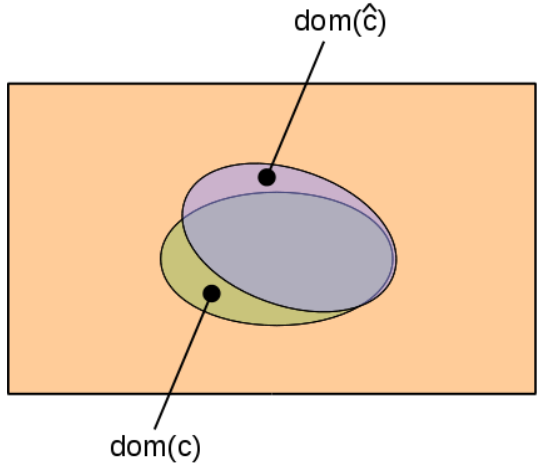}
\caption{Overlapping between the domain of a category $\ok[]$ and the domain of the corresponding classifier $\ck[]$: the more overlapping, the better the behavior of the classifier~is.} \label{fig:overlapping} 
\end{figure}

The definition of domain can also be given for classifiers. In particular, $dom(\ck[])$ denotes the set of inputs accepted by $\ck[]$. 
In the ideal case in which $dom(\ck[]) \equiv dom(\ok[])$, we say that the classifier acts as an \emph{oracle} for the given category.
However, although a classifier is expected to approximate as much as possible the corresponding category, its domain typically does not coincide with that identified by the oracle (see Figure~\ref{fig:overlapping}), i.e., $dom(\ck[]) \ne dom(\ok[])$.

\begin{figure}
\centering
\includegraphics[width=0.90\textwidth]{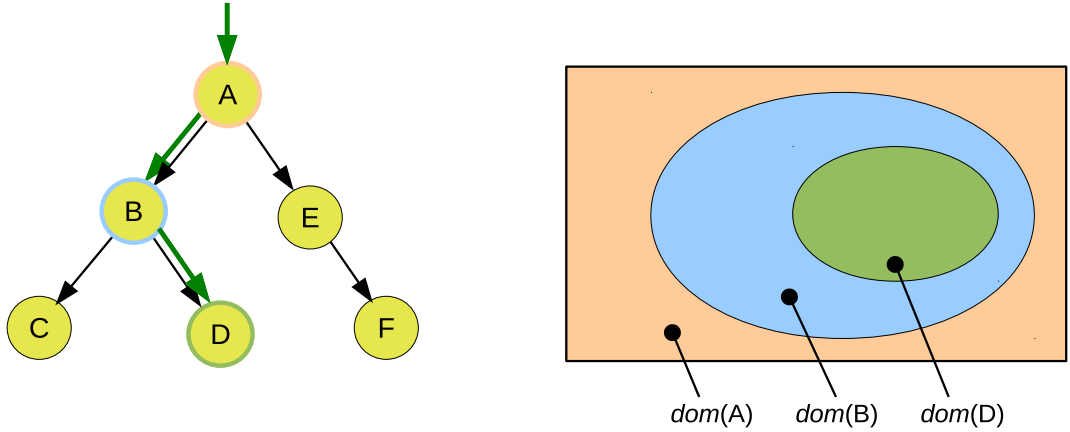}
\caption{Graphical representation of the covering relation that holds among the domains of categories occurring along a path, e.g., from A to D through B.} \label{fig:domains}
\end{figure}

Without loss of generality, we assume that the given taxonomy $\taxonomy$ has a unique root.
In principle, the domains of categories that occur along a path originating from the root satisfy an inclusion relation (see Figure~\ref{fig:domains}). The same kind of inclusion relation holds among the domains of the corresponding classifiers.

\subsection{Non-Standard Definitions for HTC}

\begin{figure}[ht]
    \centering
    \includegraphics[width=0.8\columnwidth]{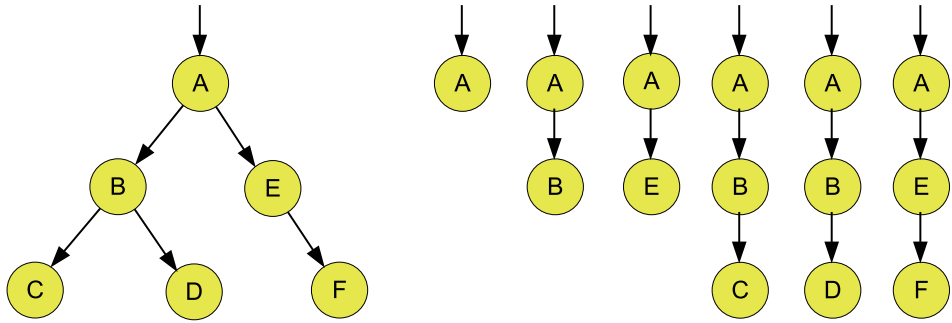}
    \caption{An example of taxonomy and its corresponding unfolding.}
    \label{fig:taxonomy-with-pipelines}
\end{figure}

We want $\emh{T}$ to be represented by the set of all its most representative paths, i.e., those that originate from the root. Any one of these paths will be called \emph{pipeline} hereinafter. Figure~\ref{fig:taxonomy-with-pipelines} depicts a simple source taxonomy, on the left part, and its ``unfolding'' in terms of pipelines, on the right part.

\begin{definition}[Well-Formed Strings and Pipelines] \label{def:pipeline}
Given a taxonomy $\taxonomy$, a pipeline $\pi$ is a well-formed string that originates from the root. 
\end{definition} \down

The definition of pipeline relies upon the concept of \emph{well-formed string}, which in turn can be defined through the corresponding characteristic function $F:\kleenestar{C} \rightarrow [0,1]$: 

\ALIGN { \label{eq:char-F}
	 F(\wfs) &=
	 \left\{ \begin{array}{ll}
	1 & \quad \wfs \equiv \lambda \vee \wfs \in \emh{C}\\
	F( \alpha) \cdot f(\beta,\alpha) \cdot F(\beta) & \quad \wfs = \alpha + \beta \equiv \alpha \beta, \; \; \alpha, \beta \in \kleenestarP{C}
	\end{array}
	\right.
}

\noindent where:
\ITEMIZE{
\item the operator ``+'' denotes concatenation between two strings (it can be omitted in absence of ambiguity);
\item the constant $\lambda$ denotes the empty string (with the property that $\forall \alpha \in \kleenestar{C}: \alpha + \lambda \equiv \lambda + \alpha \equiv \alpha$);
\item $f(\beta,\alpha)$ extends the characteristic function of the covering relation to pairs of strings in $\kleenestarP{C}$, as follows: $f(\beta,\alpha) \equiv f(head(\beta),tail(\alpha))$, with \emph{head} and \emph{tail} having the usual semantic of extracting the first and the last element of their argument, respectively.
}

Note that, in a probabilistic setting, the characteristic function $F$ represents the probability that a document will go through the corresponding pipeline under the assumption that the embedded classifiers act as oracles.

\down

The set of well-formed strings $\emh{W}_T$ and the set of pipelines $\emh{P}_T$ in $\taxonomy$ can now be defined as follows:
\ALIGN {
    \emh{W}_T &= \left\{ \wfs \in \kleenestar{C} \; \vert \; F(\wfs) > 0 \right\} \label{eq:wffs} \\
    \emh{P}_T &= \left\{ \pi \in \emh{W}_T - \{ \lambda \} \; \vert \; head(\pi) = root(\emh{T}) \right\} \label{eq:pipelines}
}

\down

For instance, the path $A \rightarrow B \rightarrow D$ shown in Figure~\ref{fig:domains} gives rise to the string $ABD$, which is well-formed, as $D \prec B \prec A$, and rooted, as $head(ABD) = root(\emh{T})$; hence, it is a pipeline. A document of category \emph{D} is expected to go through the pipeline $ABD$ if correctly classified by the corresponding taxonomic system built upon $\emh{T}$. 

\down

A partial order also holds for pipelines. The concept we want to capture here is that an existing pipeline typically embeds other pipelines. In symbols:

\up

\ALIGN{
   \pi_1, \pi_2 \in \emh{P_T}: \pi_1 \le \pi_2 \iff \exists w \in \emh{W_T} \textit{ s.t. }  \pi_2 = \pi_1 + w
}

\down

Referring once again to Figure~\ref{fig:taxonomy-with-pipelines}, other than the trivial assertion $ABC \le ABC$, we can also state that $AB \le ABC$ and that $A \le ABC$ as $AB + C = A + BC = ABC$.

The reason why pipelines are considered so important lies in the fact that they facilitate the task of analyzing the corresponding taxonomy.
In particular, pipelines can be extracted from both trees and DAGs, are immune from the problem of having to deal with multiple class labels, admit overlapping between class domains, and are naturally suited to deal with partial paths (a partial path originates from the root and terminates with an internal node of the taxonomy as opposed to a full path, which terminates with a leaf).
Consequently, all issues that may arise depending on the characteristics of a hierarchical problem require the activation of suitable policies only at the moment of moving from a pipeline-oriented to a taxonomy-oriented perspective, whereas the unfolding in terms of pipelines appears to be a common task for a variety of actual policies.

It is worth noting that the possibility of activating multiple paths within a taxonomy implies that, for at least one internal node $c \in \emh{C}$, an overlap occurs between (at least) two of its children. In symbols:
\ALIGN{
\exists \, a,b \in \emh{H}(c): \; dom(a) \cap dom(b) \neq \emptyset
}

As for the assumption of having to deal with partial paths, this implies that at least one internal node $\ok[] \in \emh{C}$ has proper instances, not shared with any of its children. In symbols:
\ALIGN{
dom(\ok[]) \supset \bigcup_{a \in \emh{H}(\ok[])} dom(a)
}

A further definition will be useful when discussing the main characteristics of PF. 
Assuming that $a, b \in \emh{C}$ and that $a \le b$, the key concept we want to capture is that the relevant inputs for $a$ with respect to $b$ are all positive instances of $b$.
To this end, let us define the notion of ``relevance set'' (\emph{rset}) between two categories, as follows:

\begin{definition}[Relevance Set] \label{def:relevance-set} 

\ALIGN{
   \forall a,b \in \emh{C}: rset(b,a) \mathdef
\begin{cases}
dom(a) \quad \text{if} \; b \le a \\
\emptyset \quad \quad \quad \; \; \text{otherwise}
\end{cases}
}

\end{definition} \down

\down

\noindent We can now check whether an input $i$ is \emph{relevant} (\emph{rel}) for a category $b$, with respect to a category $a$, according to the definition below:

\begin{definition}[Relevance] \label{def:relevance} 

\ALIGN{ rel(i, b \vert a) \mathdef i \in rset(b,a) }

\end{definition} \down

\down

\section{Progressive Filtering} \label{sec:progressive-filtering} \milestone{ANALYSIS of PF}

PF is a top-down strategy that requires the underlying taxonomy to be mirrored with local classifiers per node (hereinafter $LCN$), the underlying assumption being that the domain of a node/classifier encompasses the domains of its children. Consequently, classifiers are trained with hierarchical training sets and propagate an input only in the event that they accept it.\seenote
This choice, together with the ``pass-down'' strategy, preserves hierarchical consistency, which imposes that all ancestors of a category that accepts an input must also accept it.
As for the hierarchical consistency requirement, it may be satisfied or not, depending on the kind of structure in question (tree or DAG) and on the policy adopted to deal with well-known issues, such as: (i) leaf node prediction, (ii) premature blocking, and (iii) high-level error recovering.
We know that leaf-node prediction can be mandatory or not. In the latter case, for at least one input, its most specific class is not required to be a leaf node in the taxonomy, e.g., \cite{bib:sun01}.
When the classification stops at an internal node while an oracle would keep propagating the current input downwards, then the blocking problem arises. Some strategies to avoid blocking are discussed, for instance, in \cite{bib:sun04}.
The research community has also devoted efforts to cope with high-level error recovering. The interested reader will find several proposals aimed at tackling this issue in \cite{bib:cheng01} and in \cite{bib:wibowo02b}.
\note{A dual strategy, not considered for PF, would assume that the domain of a node/classifier accounts only for its own inputs, disregarding the domains of its children. In this case, classifiers should be trained with proper training sets and propagate an input only in the event that they do not accept it.}

\subsection{Common solutions and known issues for PF} \label{sec:known-issues}

A common solution for implementing a binary classifier $\ck[]$ for a category $\ok[] \in \emh{C}$ consists of thresholding a real-valued classifier,  entrusted with estimating the probability that an input belongs to $\ok[]$.
In so doing, an optimization problem arises, which consists of identifying the threshold that maximizes/minimizes a utility/cost function, usually a well-known metric.
The simplest solution to this problem, when classifiers are framed in a taxonomy, consists of independently optimizing pipelines of binary local classifiers, in which the \emph{same} classifier is allowed to have different thresholds, depending on which pipeline it is embedded by, \cite{bib:addis10KDIR}.
In so doing, a sort of ``flattening'' of the underlying taxonomy is performed, while pipelines still embed information about the underlying taxonomy.

PF can give rise to many other kinds of actual systems, 
depending on the given class of problems, on the choices made by the algorithm devised to solve them, and on the specific policies adopted to deal with the most well-known issues (see, for instance, \cite{bib:bennett09}) encountered while trying to enforce the hierarchical consistency requirement.

It is worth noting that even simple scenarios may hide subtle issues. Just to give a taste of them, let us consider a case in which the given problem requires mandatory leaf-prediction. This implies, at least in principle, that any input accepted by an internal node $\ok[]$ must be accepted by at least one of its children. As, by default, PF does not perform any direct action designed to enforce this property, the blocking problem may occur. 
Things deteriorate when one assumes that non-mandatory leaf-prediction is permitted, as nothing guarantees that stopping the acceptance of the current input at an internal node corresponds in fact to a correct categorization.

Summing up, a complete scenario of the probabilistic behavior of a taxonomy as a whole cannot be developed because of the large number of variations in terms of feasible policies and of the issues that may arise when trying to cope with the most well-known problems ensuing from a top-down strategy based on LCNs.
Any specific solution (with its pros and cons) would generate a different statistical behavior, though still based on the pipelines extracted from the given taxonomy. This is the main reason why we concentrate on pipelines, which allow to perform a preliminary analysis regardless of the combination policy adopted.
In particular, the focus will be first on classifiers in isolation and then on pipelines of classifiers. In both cases, the concept of ``normalized'' confusion matrix is used to differentiate the probabilistic behavior of a classifier from the actual confusion matrices that summarize the results of specific experiments.

In the following, we also assume that the behavior of all classifiers is statistically significant. Under this assumption, we can model the outcome of a classifier embedded by a pipeline with two random variables, ranging over 0 (\emph{false}) and 1 (\emph{true}). 
In particular, following the choice made to distinguish oracles from actual classifiers,
random variables related to oracles are denoted in plain format (e.g., $X$), whereas those related to actual classifiers have a circumflex  (e.g., $\Ck[]$). 
Joint or conditional probabilities involved in the modeling activity, e.g., $\jointk[]$ and $p(\Ck[] \vert \Xk[])$, are represented with $2 \times 2$ matrices.
Single random variables are also represented with $2 \times 2$ \emph{diagonal} matrices, exploiting the fact that $p(X) \equiv p(X,X)$.

\subsection{Analysis of a Single Classifier} \label{sub:classifier} \milestone{SINGLE CLASSIFIER}

Let us denote with $\Xi_c(\posex,\negex)$ the confusion matrix of a run in which a classifier $\ck[]$ embedded by a category $\ok[] \in \emh{C}$ is fed with $\totalex$ instances, of which \emph{p} are positive and \emph{n} negative.
Paying attention to keeping the same values for \emph{p} and \emph{n} on different runs, the joint probability $\jointk[c]$ is proportional, through $\totalex$, to the expected value of $\Xi_c(p,n)$.
In symbols:

\ALIGN{
E \left[ \Xi_c(p,n) \right] = \totalex \cdot \jointk[c]
}
Assuming statistical significance, the confusion matrix obtained from a single test (or, better, averaged over multiple tests) gives us reliable information on the performance of a classifier. Hence, we can write:

\ALIGN{ \label{eq:joint-prob-single-1}
 \Xi_c(p,n) \approx \totalex \cdot \jointk[c] = \totalex \cdot p(\Xk[c]) \cdot p(\Ck[c] \vert \Xk[c])
}

\Down

We assume that the transformation performed by $\ck[]$ can be isolated from the inputs it processes, at least from a statistical perspective.
In so doing, the confusion matrix for a given set of inputs can be written as the product between a term that accounts for the number of positive and negative instances, on the one hand, and a term that represents the expected recognition / error rate of $\ck[]$. In symbols:

\ALIGN{ \label{eq:xi}
    \Xi_c(\posex,\negex) = 
m \cdot \underbrace{\diagcmx{\nfk[c]}{\fk[c]}}_{\emh{O}(\ok[]) \approx p(\Xk[c])} \cdot \underbrace{\cmx{\gamma}}_{\Gamma(\ok[]) \approx p(\Ck[c] \vert \Xk[c])}
}

\up

\noindent where:

\ITEMIZE{
\item $\fk[c] = \posex/\totalex$ and $\nfk[c] = \negex / \totalex$ denote the percent of positive and negative instances, respectively;
\item $\gamma_{ij} \approx p(\Ck[c] = j \; \vert \; \Xk[c] = i), \; i,j = 0,1$, denote the percent of inputs that have been correctly classified ($i=j$) or misclassified ($i \neq j$) by $\ck[]$. In particular, $\gamma_{00}, \gamma_{01}, \gamma_{10}$, and $\gamma_{11}$ denote the percent of true negatives (TN), false positives (FP), false negatives (FN), and true positives (TP), respectively.
It can be easily verified that $\Gamma(c)$ is normalized row-by-row, i.e., that $\gamma_{00} + \gamma_{01} = \gamma_{10} + \gamma_{11} = 1$. For this reason, hereinafter an estimate of the conditional probability $p(\Ck[c] \vert \Xk[c])$ for a classifier $\ck[]$ embedded by a category $\ok[]$ will be called \emph{normalized confusion matrix}.
}

The separation between inputs and the intrinsic behavior of a classifier reported in Equation~\eqref{eq:xi} suggests an interpretation that recalls the concept of transfer function, where a set of inputs is applied to $\ck[]$. In fact, this could be interpreted alternatively as separating the optimal behavior of a classifier from the deterioration introduced by its actual filtering capabilities. 
In particular, $\emh{O}(c) \approx p(\Xk[c])$ represents the \emph{optimal behavior} obtainable when $\ck[]$ acts as an \emph{oracle}, whereas $\Gamma(c) \approx p(\Ck[c] \, \vert \Xk[c])$ represents the \emph{expected deterioration} caused by the actual characteristics of the classifier. 

\subsection{Analysis of a Pipeline of Classifiers} \label{sub:pipeline} \milestone{PIPELINES}

\begin{figure}[ht]
    \centering
    \includegraphics[width=0.7\columnwidth]{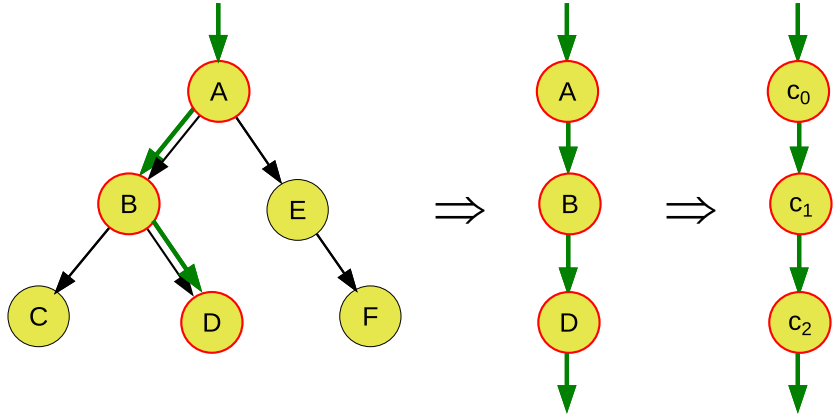}
    \caption{A pipeline of classifiers extracted from a taxonomy and undergone to standard renaming.}
    \label{fig:pipeline-from-taxonomy}
\end{figure}

Pipelines are in fact the ``building blocks" of the corresponding taxonomy.
Without loss of generality, in the following we will adopt a naming scheme independent from the generic pipeline being investigated. In particular, the components of a pipeline $\pi$ of length $L+1$ are assumed to be the categories $\ok[0], \ok[1], \ldots, \ok[L]$ (where $\ok[0]$ represents the root), the underlying assumption being that $\forall k = 1, \ldots, L: \ok[k-1] \prec \ok$. An example of pipeline, extracted from a taxonomy and undergone to standard renaming, is shown in Figure~\ref{fig:pipeline-from-taxonomy}.

Let us also assume that $\pi_k \le \pi$ denotes the ``subpipeline'' $\ok[0] \ok[1] \ldots \ok$ and that 
$e(X,\what{X})$ denotes co-occurring events involving an oracle and the corresponding classifier; in particular, $e_{ij}$ will be used as a shorthand for $e(X=i, \what{X}=j)$, $\forall i,j=0,1$.
Still for the sake of readability, the domain of $\ok$ will be denoted by $\ext{A}$, whereas the domain of $\ck$ will be denoted by $\extw{A}$.
The full list of shorthands defined with the goal of simplifying the notation while deriving relevant formulas is reported in Table~\ref{tab:shorthands}. Moreover, in absence of ambiguities, not-indexed quantities are meant to denote $k=L$, e.g., $\Omega_{\pi}(D) \equiv \Omega_{\pi_L}(D_L)$.

Studying classifiers embedded by a pipeline requires to model their interactions, which originate from the fact that the domain of a classifier $\ck$ is, by hypothesis, a proper subset of the domain of its ancestors.
While the normalized confusion matrix of a classifier $\ck[]$ in isolation originates from $p(\Ck[c] \, \vert \, \Xk[c])$, additional conditions are required for a classifier embedded by a pipeline (except for the root), which accounts for the presence of its ancestors:

\Up

\ALIGN { \label{eq:pipe-cond-prob} 
   \tok{\Gamma} \approx p(\Ck \, \vert \, \Xk, \Ck[k-1] = 1, \Ck[k-2] = 1, \ldots, \Ck[0]=1)
}

%\down

However, due to the embedding of classifiers, some tautological implications imposed by the underlying taxonomy hold for $k>0$ (see also the concept of ``True Path Rule'' in  \cite{bib:valentini11}):
\ALIGN { \label{eq:tautological-implications}
	\Xk[k-1] = 0 \models \Xk[k] = 0, \; \;  \Xk[k] = 1 \models \Xk[k-1] = 1 \quad (\text{from} \; A_{k} \subseteq A_{k-1}) \\
	\Ck[k-1] = 0 \models \Ck[k] = 0, \; \; \Ck[k] = 1 \models \Ck[k-1] = 1 \quad (\text{from} \; \what{A}_{k} \subseteq \what{A}_{k-1})
}

\down

Hence, considering that $\Ck[k-1] = 1 \models \Ck[k-2] = 1 \models \ldots \models \Ck[0] = 1$, Equation~\eqref{eq:pipe-cond-prob} can be simplified as follows:
\ALIGN { \label{eq:pipe-cond-prob-simple} 
   \tok{\Gamma} \approx p(\Ck \, \vert \, \Xk, \Ck[k-1] = 1)
}

\begin{table} \caption{Shorthands adopted while deriving relevant formulas.} \renewcommand{\arraystretch}{1.6}
\begin{tabular} {|l|l|}
\hline
\bf{Shorthand} & \bf{Explanation} \\
\hline \hline
$\pi_k = \pi_{k-1} + c_k = c_0c_1 \ldots c_k$ & Generic subpipeline ($k=0,1, \ldots, L$), $\quad \pi_k \le \pi$ \\
\hline
$\extw{A} \mathdef dom(\what{c}_k), \ext{A} \mathdef dom(c_k)$ & Domains for $\what{c}_k$ and the corresponding oracle $c_k$\\
\hline
$\Ck, \Xk$ & Random variable for $\what{c}_k$ and the corresponding oracle $c_k$ \\
\hline
$\tok{e}_{ij} \mathdef e(\Xk =i, \Ck=j)$ & Co-occurring events, with $\Xk = i$ and $\Ck = j$\\
\hline
$\fk \mathdef p(\Xk = 1 \, \vert \, \Xk[k-1] = 1), \quad \fk[0] \mathdef 1$ & Probability that an input in $A_{k-1}$ also belongs to $A_k$ \\
$\nfk \mathdef p(\Xk = 0 \, \vert \, \Xk[k-1] = 1) = 1 - \fk$ & Complement of $\fk$ \\
$\Fk \mathdef p(\Xk = 1) = \prod_{j=0}^{k}{f_{j}}$ & Probability of traversing $\pi_k$, as classifiers were oracles \\
$\nFk \mathdef p(\Xk = 0) = 1 - \Fk$ & Complement of $\Fk$ \\
\hline
$D_{k} = \left\{ \fk \vert \; j=0,1, \ldots, k \right\}$ & Set of conditional probabilities along $\pi_k$ \\
\hline
$\tok{\Gamma} \mathdef \Gamma(\ok)$ & Normalized confusion matrix of the classifier $\ck$\\
\hline
$\tok{\Xi} \mathdef \Xi_{\pi_k}(D_k;\totalex)$ & Confusion matrix for $\pi_k$, fed with $\totalex$ inputs \\
$\tok{\Omega} \mathdef \Omega_{\pi_k}(D_k)$ & Estimate of the joint probability $\jointk$ \\
$\tok{\emh{O}} \mathdef \emh{O}_{\pi_k}(D_k)$ & Estimate of the prior probability $p(\Xk)$\\
$\tok{\Phi} \mathdef \Phi_{\pi_k}(D_k)$ & Estimate of the conditional probability $p(\Ck \vert \Xk)$\\
\hline
\end{tabular} \label{tab:shorthands}
\end{table}

\subsubsection{Finding an approximation for $\jointk$} \milestone{ANALYSIS OF PIPELINES}

According to a probabilistic perspective, the starting point of our analysis is:
\ALIGN { \label{eq:joint-prob}
	E \left[ \tok{\Xi} \right] = \totalex \cdot \jointk = \totalex \cdot p(\tok{e}) 
}

As the process of estimating $\jointk$ requires approximations, let us use a specific notation for the (estimation of) the joint probability $\jointk$:

\up

\ALIGN { \label{eq:joint-prob-and-omega}
	\tok{\Omega} \approx p(\tok{e})
}

\noindent From the law of total probability, represented with the Bayes decomposition, each component of $\tok{\Omega}$ can be represented as:

\up

\ALIGN { \label{eq:omega-with-bayes-decomp}
	\tok{\omega}_{ij} \approx p(\tok{e}_{ij}) = \dsum {r,s}{}{p(\tok[k-1]{e}_{rs}) \cdot p(\tok{e}_{ij} \vert \tok[k-1]{e}_{rs}}), \quad \forall i,j = 0,1
}

For the sake of brevity, we only derive $\tok{\omega}_{00}$. The reader can consult %APPENDIX~\ref{appendix:omega}
APPENDIX~A~for further details on the derivation of $\tok{\omega}_{ij}, \; \forall i,j = 0,1$.
To keep the notation simpler, let us use ``prime'' to denote events or random variables that refer to the pipeline $\pi_k$, whereas plain text refers to $\pi_{k-1}$:

\Up

\ALIGNX {
	p(e_{00}\prm) =
	p(e_{00}) \cdot p(e_{00}\prm \vert e_{00}) +
	p(e_{01}) \cdot p(e_{00}\prm \vert e_{01}) +
	p(e_{10}) \cdot p(e_{00}\prm \vert e_{10}) +
	p(e_{11}) \cdot p(e_{00}\prm \vert e_{11})
}

\noindent where:
\ALIGNX {
p(e_{00}\prm \vert e_{00})
	  &= p(\Xk[]\prm = 0, \Ck[]\prm = 0 \, \vert \, \Xk[] = 0, \Ck[] = 0) \\
	  &= \underbrace{p(\Ck[]\prm = 0 \, \vert \, \Xk[]\prm = 0, \Xk[] = 0, \Ck[] = 0)}_{=1} \cdot \underbrace{p(\Xk[]\prm = 0 \, \vert \, \Xk[] = 0, \Ck[] = 0)}_{=1} = 1\\
p(e_{00}\prm \vert e_{01})
	  &= p(\Xk[]\prm = 0, \Ck[]\prm = 0 \, \vert \, \Xk[] = 0, \Ck[] = 1) \\
	  &= \underbrace{p(\Ck[]\prm = 0 \, \vert \, \Xk[]\prm = 0, \Xk[] = 0, \Ck[] = 1)}_{\approx \gamma_{00}\prm} \cdot \underbrace{p(\Xk[]\prm = 0 \, \vert \, \Xk[] = 0, \Ck[] = 1)}_{=1}
	    \approx \gamma_{00}\prm \\
%}
%%
%\ALIGNX {
p(e_{00}\prm \vert e_{10})
	  &= p(\Xk[]\prm = 0, \Ck[]\prm = 0 \, \vert \, \Xk[] = 1, \Ck[] = 0) \\
	  &= \underbrace{p(\Ck[]\prm = 0 \, \vert \, \Xk[]\prm = 0, \Xk[] = 1, \Ck[] = 0)}_{=1} \cdot \underbrace{p(\Xk[]\prm = 0 \, \vert \, \Xk[] = 1, \Ck[] = 0)}_{\approx \nfk[]\prm}
	  \approx \nfk[]\prm\\
p(e_{00}\prm \vert e_{11})
	  &= p(\Xk[]\prm = 0, \Ck[]\prm = 0 \, \vert \, \Xk[] = 1, \Ck[] = 1) \\
	  &= \underbrace{p(\Ck[]\prm = 0 \, \vert \, \Xk[]\prm = 0, \Xk[] = 1, \Ck[] = 1) }_{\approx \gamma_{00}\prm} \cdot \underbrace{p(\Xk[]\prm = 0 \, \vert \, \Xk[] = 1, \Ck[] = 1)}_{\approx \nfk[\prm]} 
	  \approx \gamma_{00}\prm \cdot \nfk[]\prm\\
}

\Up

\noindent Hence:
\ALIGNX {
	p(e_{00}\prm) \approx \omega_{00}\prm =  \omega_{00} + \omega_{01} \cdot \gamma_{00}\prm + \nfk[]\prm \cdot \omega_{10} + \nfk[]\prm \cdot \omega_{11} \cdot \gamma_{00}\prm \\
}

By making the derivation explicit for all $\tok{\omega}_{ij}, \; i,j =0,1$, we can approximate $\jointk$ as follows ($k>0$):

\ALIGN { \label{eq:omega-k}
	\tok{\Omega} = \left\{ \begin{array}{lllll}
	\tok{\omega}_{00} & = \; \tok[k-1]{\omega}_{00} &+ \; \tok[k-1]{\omega}_{01} \cdot \tok{\gamma}_{00} &+ \; \nfk \cdot \tok[k-1]{\omega}_{10} &+ \; \nfk \cdot \tok[k-1]{\omega}_{11} \cdot \tok{\gamma}_{00} \\ \\
	\tok{\omega}_{01} &= \; 0 &+ \; \tok[k-1]{\omega}_{01} \cdot \tok{\gamma}_{01} &+ \; 0 &+ \; \nfk \cdot \tok[k-1]{\omega}_{11} \cdot \tok{\gamma}_{01} \\ \\
	\tok{\omega}_{10} &= \; 0 &+ \; 0 &+ \; \fk \cdot \tok[k-1]{\omega}_{10} &+ \; \fk \cdot \tok[k-1]{\omega}_{11} \cdot \tok{\gamma}_{10} \\ \\
	\tok{\omega}_{11} &= \; 0 &+ \; 0 &+ \; 0 &+ \; \fk \cdot \tok[k-1]{\omega}_{11} \cdot \tok{\gamma}_{11} \\ \\
	\end{array}
	\right.
}

\down

\begin{figure}
\centering
\includegraphics[width=0.40\textwidth]{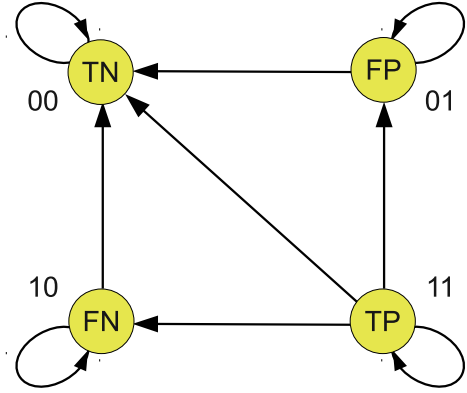}
\caption{How the elements of $\tok[k-1]{\Omega}$ concur to generate $\tok{\Omega}$. The presence of an arrow indicates that the source node (step $k-1$) contributes to the destination node (step $k$). For instance, the arrow between $TP$ and $FP$ asserts that part of $\tok[k-1]{\omega}_{11}$ is responsible for $\tok{\omega}_{01}$.} \label{fig:graph-prob-view} 
\end{figure}

To help the reader better understand the underlying process, a graphical representation of the transformation that occurs along a pipeline from step $k-1$ to step $k$ is given in Figure~\ref{fig:graph-prob-view}, which highlights how the elements of $\tok[k-1]{\Omega}$ concur to generate $\tok{\Omega}$.
Quantitative information, reported in Equation~\eqref{eq:omega-k}, is intentionally disregarded.

As for the base case (i.e., $k=0$), it can be observed that the role of the root is to forward any incoming document down to its children. In other words, the (virtual) classifier embedded by the root accepts everything as a positive instance. For this reason, the base case for $\tok[0]{\Omega}$ is:\down
\ALIGN{ \label{eq:omega-zero}
\tok[0]{\Omega} = \bmx{ 0 & 0 \\ 0 & 1}
}

\down

\noindent whereas the normalized confusion matrix of the root is:

\ALIGN{ \tok[0]{\Gamma} = \bmx{ 0 & 1 \\ 0 & 1} \mathdef \mu }

\noindent where $\neutral$ is a constant that characterizes the \emph{neutral classifier}, whose unique responsibility is to ``pass everything down'' to its children, no matter whether input documents are TP or FP.\seenote
\note{Different choices could be made to represent the normalized confusion matrix of the root, without changing the result of the transformation that occurs there. However, the adoption of the neutral classifier appears the most intuitive. We will get back to this issue in the next subsection.}

\subsubsection{Revisiting One Step of Progressive Filtering}
\begin{figure}
\centering
\includegraphics[width=0.65\textwidth]{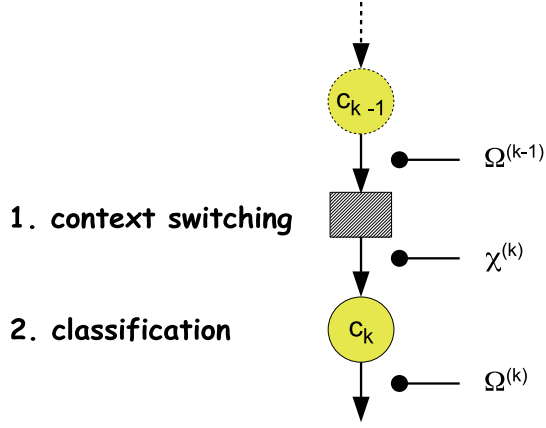}
\caption{One step of progressive filtering.} \label{fig:adaptor} 
\end{figure}

Looking at Equation~\eqref{eq:omega-k}, each processing step actually involves two separate actions. 
As sketched in Figure~\ref{fig:adaptor}, everything goes as if the output of a classifier undergo \emph{context switching} before \emph{classification}.

\down

\noindent \emph{Context switching}. Concerns the fact that only part of TP output by $\ck[k-1]$ are still TP for $\ck$. Under the assumption of statistical significance (and recalling the definition of relevance set), the percent of relevant inputs for $\ck$ that move from TP to FP is approximately $\nfk$. Conversely, only part of FN output by $\ck[k-1]$ are still FN for $\ck$, so that the percent of inputs that move from false to TN is still $\nfk$. %The underlying mechanism is sketched in Figure~\ref{fig:adaptor-mechanism}.
Hence, with $\chi$ and $\Omega$ representing the percent of inputs and the percent of outputs of a classifier in terms of true/false positives/negatives, we can write:

\ALIGN {
	\tok{\chi} & =
	\bmx{
	\tok[k-1]{\omega_{00}} + \nfk \cdot \tok[k-1]{\omega_{10}} & \tok[k-1]{\omega_{01}} + \nfk \cdot \tok[k-1]{\omega_{11}} \\
	\fk \cdot \tok[k-1]{\omega_{10}} & \fk \cdot \tok[k-1]{\omega_{11}}} =
	%\tok[k-1]{\Omega} + \nfk \cdot \bmx { 0 & +1\\ 0 & -1 } \cdot \tok[k-1]{\Omega} =
	\bmx { 1 & \nfk \\ 0 & \fk } \cdot \tok[k-1]{\Omega}
}

%\begin{figure}[b]
%\centering
%\includegraphics[width=0.45\textwidth]{./adaptor-mechanism} 
%\caption{Context switching: adapting the output of a classifier to the input of the next classifier in a pipeline.}
%\label{fig:adaptor-mechanism}
%\end{figure}

\Down

\begin{figure}
\centering
\includegraphics[width=0.50\textwidth]{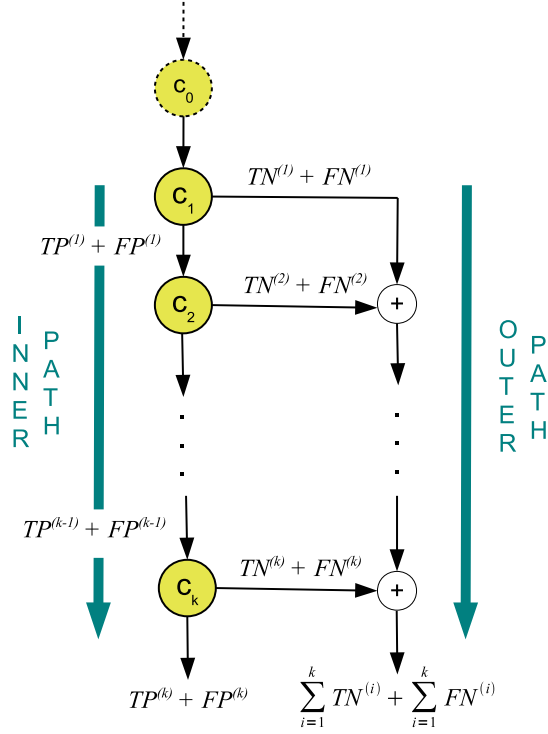} \caption{Inner and outer paths along a pipeline.} \label{fig:inner-outer-path} 
\end{figure}

\noindent \emph{Classification}. The transformation performed by $\ck$ can be better understood highlighting that two paths can be followed by a document while going through the pipeline in hand: \emph{inner} and \emph{outer} path.
Figure~\ref{fig:inner-outer-path} illustrates the different paths followed by input documents while traversing a pipeline.

\down

\noindent The \emph{inner} path operates on true positives ($\chi_{11}$) and false positives ($\chi_{01}$).
The corresponding transformation can be represented as follows:

\ALIGN { \label{eq:omega-inner}
	\tok{\Omega} \Big \vert_{inner} & =
	\innercmx {\tok{\chi_{01}} \cdot \tok{\gamma_{01}}}
			 { \tok{\chi_{11}} \cdot \tok{\gamma_{11}}}
	= \diagcmx{\tok{\chi_{01}}} {\tok{\chi_{11}}} \cdot 
	   \innercmx{\tok{\gamma_{01}}} {\tok{\gamma_{11}}}
}

\DDown

\noindent The \emph{outer} path operates on true negatives ($\chi_{00}$) and false negatives ($\chi_{10}$).
The whole process is cumulative, and can be represented as follows (still for the classifier $\ck$):

\ALIGN { \label{eq:omega-outer}
	\tok{\Omega} \Big \vert_{outer} & =
	\outercmx { \tok{\chi_{00}} + \tok{\chi_{01}} \cdot \tok{\gamma_{00}}}
			  {\tok{\chi_{10}} + \tok{\chi_{11}} \cdot \tok{\gamma_{10}}} =
	\outercmx { \tok{\chi_{00}}} {\tok{\chi_{10}}} +
	\diagcmx   {\tok{\chi_{01}}} {\tok{\chi_{11}}} \cdot
	\outercmx {\tok{\gamma_{00}}} {\tok{\gamma_{10}}} 
}

\Down

\noindent Putting together Equation~\eqref{eq:omega-inner} and \eqref{eq:omega-outer}, we obtain:
\ALIGN { \label{eq:inner-outer-path}
	\tok{\Omega}
	& =
	\outercmx { \tok{\chi_{00}}} {\tok{\chi_{10}}} +
	\diagcmx   {\tok{\chi_{01}}} {\tok{\chi_{11}}} \cdot \tok{\Gamma}
}

\Down

\noindent For its importance within the model, the transformation represented by Equation~\eqref{eq:inner-outer-path} deserves a specific definition. 

\begin{definition}[Operator $\oplus$] \label{def:oplus} 
\ALIGN {
	A \oplus B = \cmx{\alpha} \oplus \cmx{\beta}& \mathdef \Oplus {\alpha}{\beta}
}
\end{definition} \down

\noindent It is now easy to obtain a compact form for the transformation that occurs along the inner and the outer path of a pipeline. In symbols:
\ALIGN {
	\label{eq:omega}
	\tok{\Omega} =
	\tok{\chi} \oplus \tok{\Gamma} =
	\underbrace{\left( \bmx { 1 & \nfk\\ 0 & f_{k} } \cdot \tok[k-1]{\Omega} \right)}_{\text{context switching}} \oplus \underbrace{\tok{\Gamma}}_{\text{classification}}
}

\noindent Note that Equation~\eqref{eq:omega} can be applied also to the base case ($k=0$), yielding:
%\note{~Recall that, when focusing on the root: $\fk[0] = 1$, $\nfk[0] = 0$, and $\tok[0]{\Gamma} = \neutral$.}
%

\ALIGN {
	\label{eq:omega-0}
	\tok[0]{\Omega} =
	\tok[0]{\chi} \oplus \tok[0]{\Gamma} =
	\left( \bmx { 1 & \nfk[0] \\ 0 & \fk[0] } \cdot \bmx { 0 & 0 \\ 0 & 1 } \right) \oplus \tok[0]{\Gamma} =             
         \bmx { 0 & 0 \\ 0 & 1 } \oplus \neutral =  \bmx { 0 & 0 \\ 0 & 1 } 
}

\noindent Equation~\eqref{eq:omega-0} points out that neither the (virtual) context switching performed before submitting the input to the root nor the (virtual) processing of the root alter the given input --upon the assumption that $\fk[0] = 1$ (hence, $\nfk[0]=0$) and that $\tok[0]{\Gamma} = \neutral$. 

\Down

\noindent Summarizing, the overall transformation can be represented as follows: 
\ITEMIZE {
\item Base case ($k=0$), i.e., output of the root:
\ALIGN { \label{eq:omage-k-base}
	\tok[0]{\Omega}= \bmx { 0 && 0 \\ 0 && 1 }
}
\item Recursive step ($k>0$), which coincides with Equation~\eqref{eq:omega-k}:
\ALIGN { \label{eq:omega-k-step}
	\tok{\Omega}= \left( \bmx { 1 & \nfk \\ 0 & \fk } \cdot \tok[k-1]{\Omega} \right) \oplus \tok{\Gamma}
}
}

\down 

\begin{figure}
\centering
\includegraphics[width=0.80\textwidth]{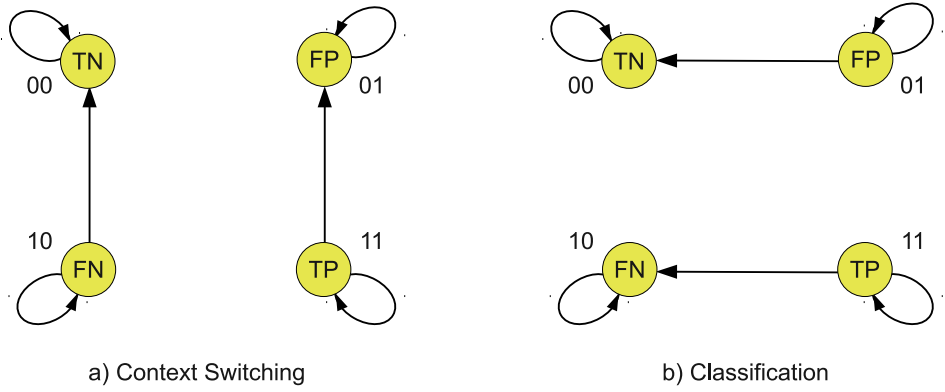}
\caption{How the elements of $\tok[k-1]{\Omega}$ concur to generate $\tok{\Omega}$, with separate focus for (a) context switching and (b) classification.} \label{fig:graph-oper-view} 
\end{figure}

Figure~\ref{fig:graph-oper-view} can help the reader better understand context switching and classification. As previously done, also in this case quantitative information is intentionally disregarded.

\Down

Unfolding the recurrence relation that defines $\Omega$ allows to obtain a closed formula, which accounts for the behavior of a pipeline $\pi_k$ ($k > 0$):

\ALIGN { \label{eq:omega-k-unfolded}
  \tok{\Omega} =
  \begin{cases}
	\tok{\omega_{00}} & =
	\nFk - \dsum{j=1}{k}{\nfk[j] \cdot \Fk[j-1]} \cdot
	\left( \dprod{r=0}{j-1}{\tok[r]{\gamma_{11}}} \right) \cdot 
	\left( {\dprod{s=j}{k}{\gamma_{01}^{(s)}}} \right) \\
	\tok{\omega_{01}} & =
	\dsum{j=1}{k}{\nfk[j] \cdot \Fk[j-1]} \cdot
	\left( \dprod{r=0}{j-1}{\tok[r]{\gamma_{11}}} \right) \cdot 
	\left( \dprod{s=j}{k}{\tok[s]{\gamma_{01}}} \right) \\
	\tok{\omega_{10}} & =
	\Fk - \Fk \cdot \dprod{j=0}{k}{\tok[j]{\gamma_{11}}} \\
	\tok{\omega_{11}} & =
	\Fk \cdot
	\dprod{j=0}{k}{\tok[j]{\gamma_{11}}} \\
  \end{cases}
}

It is easy to verify from Equation~\eqref{eq:omega-k-unfolded} that $\tok{\omega}_{00} = \nFk - \tok{\omega}_{01}$ and that $\tok{\omega}_{10} = \Fk -  \tok{\omega}_{11}$; hence let us spend a few words to clarify the underlying semantics only for $\tok{\omega}_{01}$ and $\tok{\omega}_{11}$.

As for $\tok{\omega}_{11}$, it represents the core behavior of PF. 
In particular, given an input, each classifier along the pipeline $\pi_{k}$ accepts and forwards it with probability $\fk[j] \cdot \tok[j]{\gamma_{11}}, \; j=0,1, \ldots, k$. 
The resulting product can be split in two terms, one that accounts for the distribution of inputs and the other that accounts for the intrinsic properties of classifiers, as follows:

\ALIGN{ \label{eq:omega-k-tp}
      \dprod{j=0}{k}{\left( \fk[j] \cdot \tok[j]{\gamma_{11}} \right)} =
      \left( \dprod{j=0}{k}{\fk[j]} \right) \cdot \left( \dprod{j=0}{k}{\tok[j]{\gamma_{11}}} \right) =
      \Fk \cdot \left( \dprod{j=0}{k}{\tok[j]{\gamma_{11}}} \right)
}

\down

As for $\tok{\omega}_{01}$, each component of the sum denotes a different subset of inputs, recognized as positive by $\ck$ but in fact negative. As these subsets are independent from each other, let us concentrate on a generic $j$-th element of the sum. In symbols:

\ALIGN{ \label{eq:omega-k-fp}
\nfk[j] \cdot \Fk[j-1] \cdot
	\left( \dprod{r=0}{j-1}{\tok[r]{\gamma_{11}}} \right) \cdot 
	\left( {\dprod{s=j}{k}{\gamma_{01}^{(s)}}} \right) \quad \quad j= 1, 2, \ldots, k
}

Two processing modes hold along the pipeline $\pi_k$, and the switching occurs between $\ck[j-1]$ and $\ck[j]$. Let us analyze these modes, together with the corresponding context switching:
\begin{enumerate}[(a)]
\item \emph{Processing mode before $\ck[j]$}. This behavior reproduces the one already analyzed for $\tok{\omega}_{11}$, with the obvious difference that it is observed along the pipeline $\pi_{j-1}$;
\item \emph{Context switching between $\ck[j-1]$ and $\ck[j]$}. The effect of context switching is to turn TP into FP, with probability $\nfk[j]$;
\item \emph{Processing mode after $\ck[j-1]$}. To keep ``surviving'' as FP, an input must be (incorrectly) recognized as positive by all the remaining classifiers that occur along the pipeline, including $\ck[j]$, each with probability $\tok[s]{\gamma}_{01}, \; s=j, j+1, \ldots k$.
\end{enumerate}

According to the ordering followed by the enumeration above, Equation~\eqref{eq:omega-k-fp} can be rewritten as: 

\ALIGN{ \label{eq:omega-k-fp-1}
      \underbrace{\Fk[j-1] \cdot \left( \dprod{r=0}{j-1}{\tok[r]{\gamma_{11}}} \right)}_{(a)}
      \cdot 
      \underbrace{\vphantom{\dsum{0}{1}{}}\nfk[j]}_{(b)}
      \cdot
      \underbrace{\left( \dprod{s=j}{k}{\tok[s]{\gamma_{01}}} \right)}_{(c)}  \quad \quad j= 1, 2, \ldots, k
}

Now that the semantics of all elements reported in $\tok{\Omega}$ has been clarified, let us try to give $\tok{\Omega}$ a more concise form through the following definitions: %(recall that $\tok[0]{\gamma_{01}} = \tok[0]{\gamma_{11}} = 1$, as $\tok[0]{\Gamma} = \neutral$):

\ALIGN { \label{eq:simplifying-defs}
	{\dprod{i=0}{k}{\gamma_{01}^{(i)}}} \mathdef \tok[k]{\psi_{01}}, \quad
	{\dprod{i=0}{k}{\gamma_{11}^{(i)}}} \mathdef \tok[k]{\psi_{11}}, \; \; \text{and} \; \; 
	\dfrac{1}{\nFk[k]} \cdot {\dsum{i=1}{k}{\nfk[i] \cdot \Fk[i-1]}} \cdot \dfrac{\psi_{11}^{(i-1)}}{\psi_{01}^{(i-1)}} \mathdef \tok[k]{\eta}
}

\noindent According to these definitions, $\tok{\Omega}$ can be rewritten as ($k> 0$):

\ALIGN { \label{eq:omega-k-1}
\tok{\Omega}
= \underbrace{\diagcmx { \nFk } {  \Fk }}_{\tok{\emh{O}}}
\cdot \underbrace{\bmx { 1 - \tok{\eta} \cdot \tok{\psi_{01}} & & \tok{\eta} \cdot \tok{\psi_{01}} \\ 1 - \tok{\psi_{11}} & & \tok{\psi_{11}} }}_{\tok{\Phi}}
}

\noindent where $\tok{\emh{O}}$ accounts for the optimal behavior of the pipeline $\pi_k$ (as all its classifiers were oracles), whereas $\tok{\Phi}$ represents the expected deterioration, due to the actual behavior of~$\pi_k$.

It is easy to verify that $\Phi$, which plays for pipelines the role that $\Gamma$ plays for single classifiers, is also normalized row-by-row for each $k = 1, 2, \ldots, L$.
As for $\tok[0]{\Phi}$, we know that the following equivalence must hold: $\tok[0]{\Phi} = \tok[0]{\Gamma} = \mu$. Hence, as expected, also $\tok[0]{\Phi}$ is normalized row-by-row.

Moreover, as $\tok{\Omega}$ spans over the whole space of events, the sum over its components must be 1. While trivially true for $k=0$, it is easy to show it for any $k>0$. Starting from Equation~\eqref{eq:omega-k-1}, we can write:

\ALIGN{ \label{eq:summing-omega}
\dsum{ij}{}{\omega_{ij}} 
= \nFk \cdot \tok{\phi_{00}}
+ \nFk \cdot \tok{\phi_{01}}
+ \Fk \cdot \tok{\phi_{11}}
+ \Fk \cdot \tok{\phi_{11}}
= \nFk + \Fk = 1
}

Let us also note that $\tok{\eta}$ is only apparently not defined when $\tok[j-1]\psi_{01} \equiv 0$, for some $j > 1$. For instance, assuming that an index $i$ exists such that $\tok[i]{\gamma_{01}} = 0$, we have $\forall j > i: \; \psi_{01}^{(j-1)} = 0$, which in turn implies that $\forall j > i: \; \tok[j]{\eta} = \infinity$. However, $\tok{\omega_{01}}$ (and thus $\tok{\omega_{00}}$) is \emph{still} defined for any $k \ge 0$, as:
\ALIGNX {
	\tok{\omega_{01}} & =
	\lim_{\gamma_{01}^{(i)} \rightarrow 0} \nFk \cdot \eta^{(k)} \cdot \psi_{01}^{(k)}
	\equiv
	{\displaystyle \sum_{j=1}^{k}{\nfk[j] \cdot F_{j-1}}} \cdot
	\psi_{11}^{(k)} \cdot {\displaystyle \prod_{s=j}^{k}{\gamma_{01}^{(s)}}}
	=
	{\displaystyle \sum_{j=i+1}^{k}{\nfk[j] \cdot F_{j-1}}} \cdot
	\psi_{11}^{(k)} \cdot {\displaystyle \prod_{s=j}^{k}{\gamma_{01}^{(s)}}}
}

Hence, the confusion matrix $\Xi$ for a pipeline of classifiers $\pi$, to which $\totalex$ inputs with known conditional probabilities $D$ (with reference to the categories involved in the pipeline) are applied, can always be represented as:

\ALIGN {
	\Xi_{\pi}(D;\totalex) =
        \totalex \cdot \Omega_{\pi}(D) =
	\totalex \cdot \underbrace{\diagcmx { \nFk[] } { \Fk[] }}_{\emh{O}_{\pi}(D)} \cdot
	\underbrace{
       \begin{bmatrix}
	1 -  {\eta} \cdot {\psi_{01}} & & {\eta} \cdot {\psi_{01}} \\
	1 -  {\psi_{11}} & & {\psi_{11}}
	\end{bmatrix} }_{\Phi_{\pi}(D)}
}

\down

It is now clear that $\Omega$ and $\Phi$ depend both on the conditional probabilities that characterize the flow of inputs along the pipeline (through $\eta$) and on the characteristics of the involved classifiers (through $\psi_{01}$ and $\psi_{11}$).
However, $\psi_{01}$ and $\psi_{11}$ depend only on the \emph{intrinsic properties} of the classifiers involved in a pipeline, and are in fact building blocks for defining $\Phi$ and $\Omega$. In the following subsection, we better analyze this issue.

\subsection{Intrinsic Properties of a Pipeline} \label{sub:psi}

A recursive definition for $\psi_{01}$ and $\psi_{11}$ (actually, for the matrix $\Psi$) can be easily given in terms of the ``$\oplus$'' operator, as follows:

\begin{definition} [Definition of $\Psi$] \label{def:psi-recursive}
\ALIGN { \label{eq:psi-recursive}
	\Psi^{(k)} =
	\begin{cases}
	\neutral & k=0 \\
	\tok[1]{\Gamma} & k = 1 \\
	\tok[k-1]{\Psi} \oplus \tok{\Gamma} \quad & k > 1 
	\end{cases}
}
\end{definition} \down

Where the choice of reporting the base case with $k=1$ has been introduced only for the sake of readability, as it is consistent with the base case with $k=0$. In symbols:
\ALIGN { \label{eq:psi-uno}
	\tok[1]{\Psi} = \tok[0]{\Psi} \oplus \tok[1]{\Gamma} = \neutral \oplus \tok[1]{\Gamma}
	=
	\bmx { 0 & 0 \\ 0 & 0 } + \bmx { 1 & 0 \\ 0 & 1 } \cdot \tok[1]{\Gamma}
	\equiv
	\tok[1]{\Gamma}
}

Note that the row-by-row normalization property is preserved also for $\Psi$, no matter how many classifiers occur in the pipeline.
We can verify it by induction from Equation~\eqref{eq:psi-recursive}: with $\Gamma^{(k)}$ normalized by definition and assuming that $\Psi^{(k-1)}$ is normalized, we only need to verify that $\Psi^{(k)}$preserves this property.
To show it, let us rewrite Equation~\eqref{eq:psi-recursive} as follows ($k>0$):

\up

\ALIGNX {
	\tok{\psi_{00}} &= \tok[k-1]{\psi_{00}} + \tok[k-1]{\psi_{01}} \cdot \tok{\gamma_{00}} & 
	\tok{\psi_{01}} &= \tok[k-1]{\psi_{01}} \cdot \tok{\gamma_{01}} \\
	\tok{\psi_{10}} &= \tok[k-1]{\psi_{10}} + \tok[k-1]{\psi_{11}} \cdot \tok{\gamma_{10}} &
	\tok{\psi_{11}} &= \tok[k-1]{\psi_{11}} \cdot \tok{\gamma_{11}}
}

\down

\noindent Summing up row-by-row:

\ALIGNX {
	\tok{\psi_{00}} + \tok{\psi_{01}} =
	\tok[k-1]{\psi_{00}} + \tok[k-1]{\psi_{01}} \cdot \left( \tok{\gamma_{00}} + \tok{\gamma_{01}} \right) =
	\tok[k-1]{\psi_{00}} + \tok[k-1]{\psi_{01}} = 1 \\
	\tok{\psi_{10}} + \tok{\psi_{11}} =
	\tok[k-1]{\psi_{10}} + \tok[k-1]{\psi_{11}} \cdot \left( \tok{\gamma_{10}} + \tok{\gamma_{11}} \right) =
	\tok[k-1]{\psi_{10}} + \tok[k-1]{\psi_{11}} = 1
}

\noindent Unfolding the definition of $\Psi^{(k)}$ and taking into account the normalization property we can write ($k \ge 0$):

\ALIGN { \label{eq:psi-unfolded}
	\Psi^{(k)} =
	\begin{bmatrix}
	 \dsum {j=1} {k} { \tok[i]{\gamma_{00}} } \cdot \dprod {i=0} {j-1} { \tok[i]{\gamma_{01}}}  & \ & \dprod {i = 0} {k} { \tok[i] {\gamma_{01}}} \\
	 \dsum {j=1} {k} { \tok[j]{\gamma_{10}} } \cdot \dprod {i=0} {j-1} { \tok[i]{\gamma_{11}}}  & \ & \dprod {i = 0} {k} { \tok[i] {\gamma_{11}}}
	\end{bmatrix} =
	\begin{bmatrix}
	  1 - \dprod {j=0} {k} { \tok{\gamma_{01}} } & & \dprod {j=0} {k} { \tok{\gamma_{01}} } \\
	  1 - \dprod {j=0} {k} { \tok{\gamma_{11}} } & & \dprod {j=0} {k} { \tok{\gamma_{11}}} 
	\end{bmatrix}
}

\down

It is worth pointing out that $\Psi$, which gives the expected result for $\psi_{01}$ and $\psi_{11}$, can be seen as a relaxed form of $\Phi$, in which --for a pipeline $\pi_k$-- all negative inputs are taken \emph{outside} the domain of $\ok[1]$ whereas all positive inputs are taken \emph{inside} the domain of $\ok$.
%
%\rc{In so doing, all classifiers but the root are entrusted with processing the \emph{same} set of positive and negative inputs}.
%
This choice can be imposed in the model of $\Omega$ by setting $0 < f_1 < 1$ and $f_j = 1, \; j=2, \ldots, k$ ($f_0$~is always equal to 1, by definition), so that $\eta^{(k)},\Fk \; \text{and} \; \nFk$ reduce to $1, f_1 \; \text{and} \; \nfk[1]$, respectively. 
This implies that no adaption is required for classifiers in the pipeline, except for $\ck[1]$. Under this restrictive hypothesis, $\Omega^{(k)}$ reduces to ($k > 0$):

\ALIGN { \label{eq:omega-k-intrinsic}
	\tok{\Omega} =
%	\begin{bmatrix}
%	\nfk[1] \cdot ( 1 - \tok{\psi_{01}} ) & & \nfk[1] \cdot \tok{\psi_{01}} \\ \fk[1] \cdot ( 1 - \tok{\psi_{11}} ) & & \fk[1] \cdot \tok{\psi_{11}} 
%	\end{bmatrix} =
	\underbrace{\diagcmx { \nfk[1] } { \fk[1] }}_{\tok{\emh{O}}} \cdot \underbrace{\bmx { 1 - \tok{\psi_{01}} & & \tok{\psi_{01}} \\ 1 - \tok{\psi_{11}} & & \tok{\psi_{11}}}}_{\tok{\Phi} \equiv \tok{\Psi}}
}

\down

%Equation~\eqref{eq:omega-k-intrinsic} points out that $\Phi$ coincides with $\Psi$ when no adaptation is required for $k>1$.
%
As Equation~\eqref{eq:omega-k-intrinsic} accounts \emph{only} for the internal structure of the corresponding pipeline, one can hypothesize that $\Psi$ is in fact a homomorphism which maps elements from $\kleenestar{C}$ to the space of normalized confusion matrices, say $\emh{M} \equiv [0,1]^{4}$. In symbols:

\up

\ALIGN {
	\Psi: \kleenestar{C} \rightarrow \emh{M}
}

\down

Indeed, given a taxonomy $\taxonomy$, it is easy to verify that $\Psi$ is a homomorphism, as:

\ITEMIZE {
\item the Kleene star of $\emh{C}$ yields in fact a monoid, closed under the  concatenation operation (denoted with ``$+$''), associative, and whose neutral element is the empty string $\lambda$;\\

\Up

\item the space $\emh{M}$ of normalized confusion matrices is also a monoid, closed under the ``$\oplus$'' operation, associative, and whose neutral element is the neutral classifier $\neutral$.
}

This is due to the fact that $\Psi$ preserves the structure, with ``$\oplus$'' and $\neutral$ playing in $\emh{M}$ the role that ``$+$'' and $\lambda$ play in $\kleenestar{C}$.
The interested reader can consult 
%APPENDIX \ref{appendix:homomorphism} 
APPENDIX~B~for further details concerning this issue.

\down

\begin{figure}
\center
\includegraphics[width=0.70\textwidth]{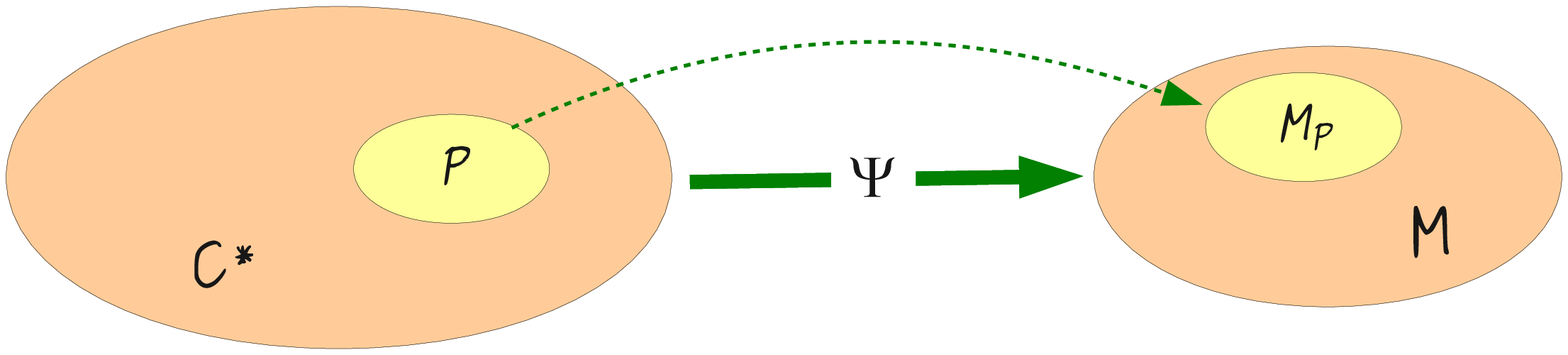} \caption{The homomorphism $\Psi$, which holds between $\kleenestar{C}$ and $\emh{M}$}
\label{fig:psi-homomorphism}
\end{figure}

Note that, although $\Psi$ is defined for any string in $\kleenestar{C}$, we are in fact interested in pipelines (see Figure~\ref{fig:psi-homomorphism}). However, as already pointed out, they can be easily identified throughout the characteristic function $F: \kleenestar{C} \rightarrow [0,1]$, which is strictly greater than zero only for well-formed strings, with the additional constraint that, to be pipelines, they must originate from the root.

Moreover, thanks to the associative property, it is also possible to give constructive definitions for $\emh{M}$ through $\Psi$. In symbols (with $c \in \emh{C}, \pi \in \kleenestar{C}, \; \text{and} \; \Psi(\lambda) \equiv \neutral$):

\ITEMIZE {
\item  \emph{Right recursion}:
$\, \pi'=\ok[]+\pi \Rightarrow \Psi(\pi')=\Psi(\ok[]+\pi)=\Psi(\ok[])\oplus\Psi(\pi)\equiv\Gamma(\ok[])\oplus\Psi(\pi)$ 
\item  \emph{Left recursion}:
$\quad \pi'=\pi+\ok[] \Rightarrow \Psi(\pi')=\Psi(\pi+\ok[])=\Psi(\pi)\oplus\Psi(\ok[])\equiv\Psi(\pi)\oplus\Gamma(\ok[])$
}

\section{Analysis on Relevant Metrics} \label{sec:metrics} \milestone{RELEVANT METRICS}

\subsection{Taxonomic Variations of Ordinary Metrics}\label{sub:metrics}

%Among various proposals aimed at evaluating the performance of hierarchical classifier systems, let us recall the metrics \emph{hP}, \emph{hR}, and \emph{hF1} defined in \cite{bib:kiritchenko06}. These metrics are in fact hierarchical variations of the well-known \emph{P}, \emph{R}, and \emph{F1}.

According to the focus of the paper, we propose straightforward definitions for precision, recall, and $F1$. To differentiate them from other proposals (e.g.,  \emph{hP}, \emph{hR}, and \emph{hF1} defined in \cite{bib:kiritchenko06}), they will be denoted as \emph{tP}, \emph{tR}, and \emph{tF1} --standing for ``taxonomic'' \emph{P}, \emph{R}, and \emph{F1}, respectively.
Their definition apply to pipelines and strictly follow the probabilistic modeling represented by the $\Omega$ matrix.
As the elements of $\tok{\Omega}$ can be used to represent the overall transformation performed by $\pi_k$, calculating \emph{tP}, \emph{tR}, and \emph{F1} for a pipeline $\pi_k$ is now straightforward:
\ALIGN {
      \label{eq:precision-pipe}
	tP(\pi_k) &= 
	\frac{\tok{\omega_{11}}} {\tok{\omega_{11}} + \tok{\omega_{01}}} =
	%\frac{\tok{\Fk \cdot \psi_{11}}} {\Fk \cdot \tok{\psi_{11}} + \nFk \cdot \tok{\psi_{01}}} =
       \left(1 + \dfrac{\tok{\omega_{01}}}{\tok{\omega_{11}}} \right)^{-1}  =
	\left(1 + \tok{\eta} \cdot \frac{\nFk}{\Fk} \cdot \dfrac{\tok{\psi_{01}}}{\tok{\psi_{11}}} \right)^{-1} \\
	\label{eq:recall-pipe}
	tR(\pi_k) &= 
	\frac{\tok{\omega_{11}}} {\tok{\omega_{11}} + \tok{\omega_{10}}} =
      \frac{\Fk \cdot \psi_{11}^{(k)}}{\Fk \cdot \tok{\psi_{11}} + \Fk \cdot \tok{\psi_{10}}} = \tok{\psi_{11}} \\
	\label{eq:F1-pipe}
	tF_{1}(\pi_{k}) &= 2 \cdot \left( \frac{1} {P} + \frac{1}{R}  \right)^{-1} = 2 \cdot \left[ \left( 1 + \tok{\eta} \cdot \frac{\nFk}{\Fk} \cdot \frac{\tok{\psi_{01}}}{\tok{\psi_{11}}} \right) + \left( \frac{1}{\tok{\psi_{11}}} \right) \right]^{-1}
%  \label{eq:accuracy-pipe}
%  tA(\pi_k) &= 
%  \frac{\tok{\omega_{00}} + \tok{\omega_{11}}} {\sum_{ij} \tok{\omega_{ij}}}
%  = \tok{\omega_{00}} + \tok{\omega_{11}} = tr(\tok{\Omega})
%  = \nFk \cdot ( 1 - \tok{\eta} \cdot \tok{\psi_{01}} ) + \Fk \cdot \tok{\psi_{11}}
}

\noindent As for the taxonomic accuracy (\emph{tA}), although less important for assessing the behavior of classifiers in pipeline (mainly due to the fact that usually the imbalance between positive and negative inputs rapidly grows with the depth level of the classifier under analysis), it can be easily defined as well:

\ALIGN{ \label{eq:accuracy-pipe}
  tA(\pi_k) &= 
  \frac{\tok{\omega_{00}} + \tok{\omega_{11}}} {\sum_{ij} \tok{\omega_{ij}}}
  %= \tok{\omega_{00}} + \tok{\omega_{11}}
  = tr(\tok{\Omega})
  = \nFk \cdot \left( 1 - \tok{\eta} \cdot \tok{\psi_{01}} \right) + \Fk \cdot \tok{\psi_{11}}
}

\noindent where $tr(\cdot)$ denotes the \emph{trace} of a matrix, obtained by summing up the elements of its main diagonal.

\subsection{How Taxonomic Precision, Recall, and $F_1$ change along a pipeline}\label{sec:metrics-change-td}

To better assess \emph{PF}, let us analyze how the most relevant metrics change along a pipeline $\pi$.
Our analysis proceeds by induction, assuming of having assessed the behavior of $\pi_{k-1}$, and then verifying what happens if one adds a further classifier.\seenote
Again for the sake of readability, let us distinguish any relevant parameter concerning the pipelines $\pi_k$ and $\pi_{k-1}$ with a ``prime'' and a ``plain'' notation, respectively (for instance, $\Fk[]\prm$ denotes $\Fk$, whereas $\Fk[]$ denotes $\Fk[k-1]$).
\note{~As, by definition, $P(\pi_0) = R(\pi_0) = F1(\pi_0) \equiv 1$, the constraints for $\pi_1$ are in fact not relevant. For this reason, the analysis concerns only pipelines $\pi_k$ with $k>1$.}

\ITEMIZE {

\item{\emph{Precision}} -- 
Imposing that $tP(\pi\prm) - tP(\pi) \geq 0$, the constraint on $tP$ that involves the relevant parameters of $\ck$ is:
\ALIGN{ \label{eq:constraint-on-precision}
	\gamma_{01}\prm \le
	\dfrac{\nFk[] \eta}{\nFk[]\prm \eta\prm} \cdot \fk[]\prm \cdot \gamma_{11}\prm
}

\up

\noindent where:
\ALIGNX{
\nFk[]\prm \eta\prm =
	{\displaystyle \sum_{j=1}^{k}{\nfk[j] \cdot F_{j-1}}} \cdot \dfrac{\tok{\psi_{11}}}{\tok[j-1]{\psi_{01}}}
	 = 
	\nFk[] \eta + \nfk[]\prm \cdot \Fk[] \cdot  \dfrac{\psi_{11}}{\psi_{01}} \gt \nFk[] \eta
}

Summarizing, $tP$ may increase or not along a pipeline depending on the constraint reported in Equation \eqref{eq:constraint-on-precision}, which is strictly related with the behavior of the ratio $\nFk[] \eta /\nFk[]\prm \eta\prm$. Note that the behavior of $tP$ depends on the distribution of data expected to flow along the pipeline.

\down

\item{\emph{Recall}} --
Imposing that $tR(\pi\prm) - tR(\pi) \geq 0$, the constraint on $tR$ that involves the relevant parameters of the classifier $\ck$ is:
\ALIGN{ \label{eq:constraint-on-recall}
\gamma_{11}\prm \ge 1
}

It is clear that the constraint on $tR$ is satisfied only when $\gamma_{11}\prm = 1$. Hence, $tR$ is \emph{monotonically decreasing} along a pipeline, the lower $\gamma_{11}\prm$ (i.e., the percent of TP) the greater the decrement of \emph{R}. Note that the behavior of $tR$ does not depend on the distribution of data expected to flow along the pipeline.

\item{\emph{$F_1$}} -- According to the given definition, $tF1$ lies in between $tP$ and $tR$. It is typically decreasing, unless the expected behavior of $tR$ (monotonically decreasing) is more than counterbalanced by an increase of $tP$.
} % end ITEMIZE

\section{Discussion} \label{sec:discussion} \milestone{DISCUSSION}

Two main questions have been formulated at the beginning of the paper (Section~\ref{sec:introduction}), concerning (i) the possibility of predicting the expected behavior of a system implemented in accordance with PF when fed with a corpus of documents whose statistical properties are known and (ii) the possibility of separating the statistical information concerning inputs from the intrinsic properties of the classifiers embedded in the given taxonomy.
After focusing on the above questions, the discussion will also summarize the analysis performed on relevant metrics. %Some real-world scenarios in which the probabilistic model proposed in this paper can be used will be also discussed.

\subsection{Predicting the Expected behavior of a PF System}

We have shown that it is very difficult to analyze a taxonomy as a whole, also due to the number of variants that can be put into practice while trying to enforce the hierarchical consistency requirement. Rather, it becomes surprisingly easy upon the extraction of the corresponding set of pipelines.

As the process of unfolding a taxonomy can be put into practice in many different scenarios, the analysis in terms of pipelines is apparently a common step for any LCN approach, including PF.
Indeed, the unfolding process does not require specific constraints to be satisfied by the problem in hand. In particular, it can be performed in presence of (i) trees or DAGs, (ii) non-overlapping or overlapping among (the domains of) categories, and (iii) mandatory or non-mandatory leaf-node prediction.

Starting from the assumption that the confusion matrix measured after performing an experiment with a pipeline $\pipe$ is in fact a single realization of a probabilistic process, the following equation holds (see \S~\ref{sub:pipeline}):

\ALIGN{  \label{eq:jointp-omega-discussion}
    \Xi_{\pi}(D;\totalex) = \totalex \cdot \Omega_{\pi}(D) \approx \totalex \cdot \jointk[L]
}

\noindent where $\Omega_{\pi}(D)$ accounts for the behavior of $\pi$ from a probabilistic perspective, as it is an estimation of the joint probability $\jointk[L]$. An \emph{effective procedure} for evaluating $\Omega$ has been given, according to the knowledge about the behavior of the classifiers embedded by the pipeline, represented by their normalized confusion matrices $\Gamma(c_k), k=0,1, \ldots, L$, and about the expected distribution of inputs. 
Summarizing, the answer to the first question is positive, as one can easily use the analysis performed in terms of pipelines to predict the behavior of a hierarchical system compliant with PF.
Note that the analysis can be performed only when the distribution of data is known, otherwise the model will not approximate well the real-world. However, for large scale data, e.g., web applications that process user queries, the hypothesis of knowing the distribution of data is not difficult to fulfill.

\subsection{Separating the Statistical Information Concerning Inputs from the Intrinsic Properties of Classifiers}

We have shown (\S~\ref{sub:pipeline}) that $\Omega_{\pi}(D)$ can be represented as the product between $\emh{O}_{\pi}(D)$ and $\Phi_{\pi}(D)$. Considering that $\Omega_{\pi}(D)$ approximates the joint probability $p(\Xk[L], \Ck[L])$, we can write:

\Up

\ALIGN{ \label{eq:jointp-discussion} 
    \jointk[L] \approx \Omega_{\pi}(D) = \emh{O}_{\pi}(D) \cdot \Phi_{\pi}(D)
}

\noindent where $\emh{O}_{\pi}(D) \approx p(\Xk[L])$ denotes the behavior of a pipeline under the hypothesis that all classifiers it embeds were acting as oracles, whereas $\Phi_{\pi}(D) \approx p(\Ck[L] \vert \Xk[L])$ represents the expected deterioration.
We have pointed out that the $\Phi$ plays for pipelines the role that $\Gamma$ plays for single classifiers. 
However, although the property of row-by-row normalization is satisfied for both $\Phi$ and $\Gamma$, $\Phi$ \emph{still} depends on the distribution of input data while $\Gamma$ does not. 
Fortunately, some building blocks have been identified, characterized by the $\Psi$ matrix, whose elements depend \emph{only} on the intrinsic properties of the pipeline.
The dependence of $\Phi$ from $\Psi$ is highlighted by the following formula, which is very important in the process of pipeline analysis:

\ALIGN{ \label{eq:omega-discussion}
  \Omega_{\pi}(D) =
  \underbrace{\diagcmx { \nFk[] } { \Fk[] }}_{\emh{O}_{\pi}(D)}
  \cdot
  \underbrace{
       \begin{bmatrix}
	1 -  {\eta} \cdot {\psi_{01}} & & {\eta} \cdot {\psi_{01}} \\
	1 - {\psi_{11}} & & {\psi_{11}}
	\end{bmatrix} }_{\Phi_{\pi}(D)}
}

Note that, due to its independence from the distribution of data, the task of calculating $\Psi$ can be done once, and requires to be repeated only in the event that the properties of one or more classifiers in the pipeline change.
Summarizing, the answer here is only partially positive, as the approximated model represented by $\Omega$ cannot be expressed in a way that clearly separates the distribution of input data (through $\emh{O}$) from the intrinsic behavior of the pipeline (through $\Phi$). In fact, $\Phi$ still embeds the information about the distribution of input data. However, one can calculate $\Psi(\pi)$ for each pipeline $\pi \in \emh{P_T}$, starting from the normalized confusion matrices $\Gamma$ of the classifiers embedded by $\pi$. The $\Psi$ matrices are independent from the labeling of the input data, and can be used, together with the set of conditional probabilities that characterize the inclusion relations for the given pipeline, to calculate the normalized confusion matrix of the pipeline (i.e., $\Phi$) and therefore the approximated model (i.e., $\Omega$).

% NEW SUBSECTION (ML testing)

As a noticeable consequence of Equation~\eqref{eq:omega-discussion}, testing the behavior of a pipeline for a specific value of imbalance is not straightforward.
The motivation lies in the fact that the same imbalance can be obtained with many different distributions of inputs. To better highlight this issue, let us assume that one wants to measure the behavior of a pipeline in presence of 10\% of positive vs. 90\% of negative inputs.
Positive inputs refer to the last classifier in the pipeline, and their amount is fixed (in this case, 10\%).
On the other hand, negative inputs can be selected in a variety of ways along the pipeline. For instance, one may select only inputs that do not belong to any category but the root (which by hypothesis accepts all inputs).\seenote
Another peculiar policy may consist of selecting as negative inputs only those that belong to the last but one classifier in the pipeline.
However, the above policies for negative input selection are not representative enough for identifying the behavior of a pipeline in presence of imbalance. In fact, many other selection policies are feasible, provided that the constraint on imbalance is satisfied.
Summarizing, while the problem of setting up test beds with statistical significance remains (no matter whether the corresponding tests are performed with a single run, averaging over multiple runs, or resorting to \emph{k}-fold cross validation), a further problem arises for pipeline testing, as its behavior depends on the distribution of inputs being processed. Hence, studying the imbalance requires at least an averaging over multiple test, each run with a different distribution of (negative) inputs.
\note{This issue has already been described in \S~\ref{sub:psi}, while defining the $\Psi$ matrix.} %

\subsection{Analysis Performed on Relevant Metrics}

The analysis performed on relevant metrics (i.e., $tP$, $tR$, $tF_1$) highlights that $tP$ depends on the distribution of data while $tR$ does not. As for $tR$, we have shown that it is monotonically decreasing.
This phenomenon is related with the problem of high-level error recovering, which originates from the fact that errors made at higher levels of a taxonomy have great impact on the overall performance of any actual system that implements top-down processing (including those based on PF). 
The impact of high-level errors on the overall performance of a system can be better understood recalling the concepts of inner and outer path: the former is entrusted with performing progressive filtering, whereas the latter accumulates inputs that have been rejected by any of the classifiers embedded by the pipeline.
For this reason, there is no way to recover errors performed along the outer path (FN), while errors performed by a classifier along the inner path (FP) may be recovered by subsequent processing steps. This behavior is also highlighted by the study made on relevant metrics, where the recall (related to FN) is monotonically decreasing, whereas the precision (related to FP) may be increasing or not depending on the characteristics of the involved classifiers.
A simple strategy headed to limit the impact of high-level errors can be put into practice by lowering the thresholds of the embedded classifiers, the closer the classifier to the root, the lower the threshold. In so doing, FN are expected to decrease while FP are expected to increase. However, FP can be further processed by the  classifiers that occur after the current one in the given pipeline, thus literally realizing ``progressive filtering''. This strategy affects also the training of classifiers, which are required to maintain the same discrimination capabilities on relevant and non relevant inputs that originate from their ancestors (see the definition of relevant input given in Section~\ref{sec:htc}).
The main consequence of relaxing the behavior of $\ck[k-1]$ (more generally, of the pipeline $\pi_{k-1}$) is that the set of relevant inputs for $\ck$ is extended with FP that originate from $\ck[k-1]$ and its ancestors. Hence, the training activity should be performed taking into account this issue, with the goal of improving the robustness of $\ck$ towards non-relevant FP.

\section{Conclusions and Future Work} \label{sec:conclusions} \milestone{CONCLUSIONS}

In this paper, a formal modeling of the progressive filtering technique has been performed, according to a probabilistic perspective and framed within the research field of hierarchical text categorization. 
In particular, the focus has been on how to assess pipelines extracted from a  given taxonomy.
Specific care has been taken in identifying whether some building blocks exist in the model, which are independent from the underlying distribution of input data. This part of the analysis has brought to the definition of the $\Psi$ matrix, which accounts only for the structural aspects of a pipeline. How to separate the expected optimal behavior of a pipeline from the deterioration introduced by the actual classifiers it embeds is another important result. 
The way relevant metrics change along a pipeline has also been investigated. As expected, the precision may increase or decrease depending on the characteristics of the embedded classifiers, whereas the recall is monotonically decreasing along a pipeline. To limit the impact of this latter unwanted behavior, one may relax the behavior of classifiers at higher levels, thus reducing the overall percent of FN.
The results of the analysis performed in this paper should facilitate the designer of a system based on progressive filtering in the task of devising, training and testing it. In particular, some relevant scenarios have been sketched in Section~\ref{sec:introduction}, in which the proposed probabilistic model can be useful. 

As for future work, we are currently investigating the problem of which policy should be applied to train classifiers embedded in a taxonomy. Moreover, we are about to use the model in a problem of threshold optimization.

 \milestone{ACKNOWLEDGEMENTS}

\begin{acknowledgements}
Many thanks to all people that gave me help in the task of conceptualizing and formalizing 
this work. Special thanks go to Eloisa Vargiu, who first conjectured the possibility that
part of the FN disregarded by a classifier should be turned into TN, and to 
Mauro Parodi, for his wealth of ideas about linear transformations
and their application to model text categorization tasks in a hierarchical setting.
%; and to Hector Freytes, for his help in formalizing the homomorphism between pipelines and normalized confusion matrices.
Many thanks also to Donato Malerba, Giorgio Valentini, and Fabio Roli, who made a preliminary review of the manuscript.
\end{acknowledgements}

%\nocite{*}
\bibliographystyle{fundam}
\bibliography{references}

%%% APPENDICES %%%

\newpage 

\appendix

\milestone{APPENDIX A.}

\section*{Appendix A. Estimate of the Joint Probability $\jointk$}

\setcounter{section}{0}

%\input{armano-appendix-A}

%\section{Estimating the Joint probability $\jointk$} \label{appendix:omega}

From the law of total probability, we can represent $p(\Xk=i, \Ck=j) \equiv p(\tok{e_{ij}})$ as follows:
\ALIGNX {
	p(\tok{e_{ij}}) = \dsum{r,s}{}{p(\tok[k-1]{e}_{rs}) \cdot p(\tok{e}_{ij} \, \vert \, \tok[k-1]{e}_{rs})}
}

Our goal is to derive an approximated model for $p(\tok{e_{ij}})$. To differentiate between the actual probability and its approximation, the latter is denoted with $\tok{\omega}_{ij}$. 

\down

For the sake of readability, we use ``prime'' to denote events or random variables that refer to a pipeline $\pi_k = \pi_{k-1} + c_k$, whereas plain text is used for $\pi_{k-1}$. 
Before deriving the estimation of the joint probability, let us recall that the following tautological implications hold:

\up

\ALIGNX {
	\Xk[] = 0 \models \Xk[]\prm = 0, \; \;  \Xk[]\prm = 1 \models \Xk[] = 1 \\
	\Ck[] = 0 \models \Ck[]\prm = 0, \; \; \Ck[]\prm = 1 \models \Ck[] = 1
}

\down

% OMEGA_00\prm

\noindent \fbox{\emph{Estimate of} $p(e_{00}\prm)$}
\ALIGNX {
       \vartriangleright \; \; 
	p(e_{00}\prm) &=
	p(e_{00}) \cdot p(e_{00}\prm \vert e_{00}) +
	p(e_{01}) \cdot p(e_{00}\prm \vert e_{01}) +
	p(e_{10}) \cdot p(e_{00}\prm \vert e_{10}) +
	p(e_{11}) \cdot p(e_{00}\prm \vert e_{11})
}
\noindent where:
\ALIGNX {
p(e_{00}\prm \vert e_{00})
	  &= p(\Xk[]\prm = 0, \Ck[]\prm = 0 \, \vert \, \Xk[] = 0, \Ck[] = 0) \\
	  &= \underbrace{p(\Ck[]\prm = 0 \, \vert \, \Xk[]\prm = 0, \Xk[] = 0, \Ck[] = 0)}_{=1} \cdot \underbrace{p(\Xk[]\prm = 0 \, \vert \, \Xk[] = 0, \Ck[] = 0)}_{=1} = 1\\
p(e_{00}\prm \vert e_{01})
	  &= p(\Xk[]\prm = 0, \Ck[]\prm = 0 \, \vert \, \Xk[] = 0, \Ck[] = 1) \\
	  &= \underbrace{p(\Ck[]\prm = 0 \, \vert \, \Xk[]\prm = 0, \Xk[] = 0, \Ck[] = 1)}_{\approx \gamma_{00}\prm} \cdot \underbrace{p(\Xk[]\prm = 0 \, \vert \, \Xk[] = 0, \Ck[] = 1)}_{=1}
	    \approx \gamma_{00}\prm\\
p(e_{00}\prm \vert e_{10})
	  &= p(\Xk[]\prm = 0, \Ck[]\prm = 0 \, \vert \, \Xk[] = 1, \Ck[] = 0) \\
	  &= \underbrace{p(\Ck[]\prm = 0 \, \vert \, \Xk[]\prm = 0, \Xk[] = 1, \Ck[] = 0)}_{=1} \cdot \underbrace{p(\Xk[]\prm = 0 \, \vert \, \Xk[] = 1, \Ck[] = 0)}_{\approx \nfk[]\prm}
	  \approx \nfk[]\prm\\
p(e_{00}\prm \vert e_{11})
	  &= p(\Xk[]\prm = 0, \Ck[]\prm = 0 \, \vert \, \Xk[] = 1, \Ck[] = 1) \\
	  &= \underbrace{p(\Ck[]\prm = 0 \, \vert \, \Xk[]\prm = 0, \Xk[] = 1, \Ck[] = 1)}_{\approx \gamma_{00}\prm} \cdot \underbrace{p(\Xk[]\prm = 0 \, \vert \, \Xk[] = 1, \Ck[] = 1)}_{\approx \nfk[]\prm}
	  \approx \gamma_{00}\prm \cdot \nfk[]\prm
}

\noindent Hence:
\ALIGNX {
	p(e_{00}\prm) \approx \omega_{00}\prm =  \omega_{00} + \gamma_{00}\prm \cdot \omega_{01} + \nfk[]\prm \cdot \omega_{10} + \nfk[]\prm \cdot \omega_{11} \cdot \gamma_{00}\prm \\
}

% OMEGA_01\prm

\noindent \fbox{\emph{Estimate of} $p(e_{01}\prm)$}

\ALIGNX {
       \vartriangleright \; \; 
	p(e_{01}\prm) &=
	p(e_{00}) \cdot p(e_{01}\prm \vert e_{00}) +
	p(e_{01}) \cdot p(e_{01}\prm \vert e_{01}) +
	p(e_{10}) \cdot p(e_{01}\prm \vert e_{10}) +
	p(e_{11}) \cdot p(e_{01}\prm \vert e_{11})
}

\noindent where:

\ALIGNX {
p(e_{01}\prm \vert e_{00})
	  &= p(\Xk[]\prm = 0, \Ck[]\prm = 1 \, \vert \, \Xk[] = 0, \Ck[] = 0)\\
	  &= \underbrace{p(\Ck[]\prm = 1 \, \vert \, \Xk[]\prm = 0, \Xk[] = 0, \Ck[] = 0)}_{=0} \cdot p(\Xk[]\prm = 0 \, \vert \, \Xk[] = 0, \Ck[] = 0) = 0\\
p(e_{01}\prm \vert e_{01})
	  &= p(\Xk[]\prm = 0, \Ck[]\prm = 1 \, \vert \, \Xk[] = 0, \Ck[] = 1) \\
	  &= \underbrace{p(\Ck[]\prm = 1 \, \vert \, \Xk[]\prm = 0, \Xk[] = 0, \Ck[] = 1)}_{\approx \gamma_{01}\prm} \cdot \underbrace{p(\Xk[]\prm = 0 \, \vert \, \Xk[] = 0, \Ck[] = 1)}_{=1}
	    \approx \gamma_{01}\prm\\
p(e_{01}\prm \vert e_{10})
	  &= p(\Xk[]\prm = 0, \Ck[]\prm = 1 \, \vert \, \Xk[] = 1, \Ck[] = 0) \\
	  &= \underbrace{p(\Ck[]\prm = 1 \, \vert \, \Xk[]\prm = 0, \Xk[] = 1, \Ck[] = 0)}_{=0} \cdot p(\Xk[]\prm = 0 \, \vert \, \Xk[] = 1, \Ck[] = 0) = 0\\
p(e_{01}\prm \vert e_{11})
	  &= p(\Xk[]\prm = 0, \Ck[]\prm = 1 \, \vert \, \Xk[] = 1, \Ck[] = 1) \\
	  &= \underbrace{p(\Ck[]\prm = 1 \, \vert \, \Xk[]\prm = 0, \Xk[] = 1, \Ck[] = 1)}_{\approx \gamma_{01}\prm} \cdot \underbrace{p(\Xk[]\prm = 0 \, \vert \, \Xk[] = 1, \Ck[] = 1)}_{\approx \nfk[]}
	  \approx \gamma_{01}\prm \cdot \nfk[]\prm
}

\noindent Hence:
\ALIGNX {
      p(e_{01}\prm) \approx \omega_{01}\prm & = 0 + \gamma_{01}\prm \cdot \omega_{01} + 0 + \nfk[]\prm \cdot \gamma_{01}\prm \cdot \omega_{11} \\
}

% OMEGA_10\prm

\noindent \fbox{\emph{Estimate of} $p(e_{10}\prm)$}

\ALIGNX {
       \vartriangleright \; \; 
	p(e_{10}\prm) &=
	p(e_{00}) \cdot p(e_{10}\prm \vert e_{00}) +
	p(e_{01}) \cdot p(e_{10}\prm \vert e_{01}) +
	p(e_{10}) \cdot p(e_{10}\prm \vert e_{10}) +
	p(e_{11}) \cdot p(e_{10}\prm \vert e_{11})
}

\Down

\noindent where:

\ALIGNX {
p(e_{10}\prm \vert e_{00})
	  &= p(\Xk[]\prm = 1, \Ck[]\prm = 0 \, \vert \, \Xk[] = 0, \Ck[] = 0) \\
	  &= \underbrace{p(\Ck[]\prm = 0 \, \vert \, \Xk[]\prm = 1, \Xk[] = 0, \Ck[] = 0)}_{=0} \cdot p(\Xk[]\prm = 1, \, \vert \, \Xk[] = 0, \Ck[] = 0) = 0\\
p(e_{10}\prm \vert e_{01})
	  &= p(\Xk[]\prm = 1, \Ck[]\prm = 0 \, \vert \, \Xk[] = 0, \Ck[] = 1) \\
	  &= \underbrace{p(\Ck[]\prm = 0 \, \vert \, \Xk[]\prm = 1, \Xk[] = 0, \Ck[] = 1)}_{=0} \cdot p(\Xk[]\prm = 1 \, \vert \, \Xk[] = 0, \Ck[] = 1) = 0 \\
}
\ALIGNX {
p(e_{10}\prm \vert e_{10})
	  &= p(\Xk[]\prm = 1, \Ck[]\prm = 0 \, \vert \, \Xk[] = 1, \Ck[] = 0) \\
	  &= \underbrace{p(\Ck[]\prm = 0 \, \vert \, \Xk[]\prm = 1, \Xk[] = 1, \Ck[] = 0)}_{=1} \cdot \underbrace{p(\Xk[]\prm = 1 \, \vert \, \Xk[] = 1, \Ck[] = 0)}_{\approx \fk[]\prm}
	  \approx \fk[]\prm\\
p(e_{10}\prm \vert e_{11})
	  &= p(\Xk[]\prm = 1, \Ck[]\prm = 0 \, \vert \, \Xk[] = 1, \Ck[] = 1) \\
	  &= \underbrace{p(\Ck[]\prm = 0 \, \vert \, \Xk[]\prm = 1, \Xk[] = 1, \Ck[] = 1)}_{\equiv \gamma_{10}\prm} \cdot \underbrace{p(\Xk[]\prm = 1 \, \vert \, \Xk[] = 1, \Ck[] = 1)}_{\approx \nfk[]\prm}
	  \approx \gamma_{10}\prm \cdot \nfk[]\prm
}

\noindent Hence:
\ALIGNX {
      p(e_{10}\prm) & \approx \omega_{10}\prm = 0 + 0 + \fk[]\prm \cdot \omega_{10} + \fk[]\prm \cdot \gamma_{10}\prm \cdot \omega_{11} \\
}

% OMEGA_11\prm

\noindent \fbox{\emph{Estimate of} $p(e_{11}\prm)$}

\ALIGNX {
       \vartriangleright \; \; 
	p(e_{11}\prm) &=
	p(e_{00}) \cdot p(e_{11}\prm \vert e_{00}) +
	p(e_{01}) \cdot p(e_{11}\prm \vert e_{01}) +
	p(e_{10}) \cdot p(e_{11}\prm \vert e_{10}) +
	p(e_{11}) \cdot p(e_{11}\prm \vert e_{11})
}

\down
\noindent where:

\ALIGNX {
p(e_{11}\prm \vert e_{00})
	  &= p(\Xk[]\prm = 1, \Ck[]\prm = 1 \, \vert \, \Xk[] = 0, \Ck[] = 0) \\
	  &= \underbrace{p(\Ck[]\prm = 1 \, \vert \, \Xk[]\prm = 1, \Xk[] = 0, \Ck[] = 0)}_{=0} \cdot p(\Xk[]\prm = 1 \, \vert \, \Xk[] = 0, \Ck[] = 0) = 0\\
p(e_{11}\prm \vert e_{01})
	  &= p(\Xk[]\prm = 1, \Ck[]\prm = 1 \, \vert \, \Xk[] = 0, \Ck[] = 1) \\
	  &= \underbrace{p(\Ck[]\prm = 1 \, \vert \, \Xk[]\prm = 1, \Xk[] = 0, \Ck[] = 1)}_{=0} \cdot p(\Xk[]\prm = 1 \, \vert \, \Xk[] = 0, \Ck[] = 1) = 0\\
p(e_{11}\prm \vert e_{10})
	  &= p(\Xk[]\prm = 1, \Ck[]\prm =1 \, \vert \, \Xk[] = 1, \Ck[] = 0) \\
	  &= \underbrace{p(\Ck[]\prm = 1 \, \vert \, \Xk[]\prm = 1, \Xk[] = 1, \Ck[] = 0)}_{=0} \cdot p(\Xk[]\prm = 1 \, \vert \, \Xk[] = 1, \Ck[] = 0) = 0\\
p(e_{11}\prm \vert e_{11})
	  &= p(\Xk[]\prm = 1, \Ck[]\prm = 1 \, \vert \, \Xk[] = 1, \Ck[] = 1) \\
	  &= \underbrace{p(\Ck[]\prm = 1 \, \vert \, \Xk[]\prm = 1, \Xk[] = 1, \Ck[] = 1)}_{\equiv \gamma_{11}\prm} \cdot \underbrace{p(\Xk[]\prm = 1 \, \vert \, \Xk[] = 1, \Ck[] = 1)}_{\approx \fk[]\prm}
	  \approx \gamma_{11}\prm \cdot \fk[]\prm
}

\noindent Hence:
\ALIGNX {
     p(e_{11}\prm) & \approx \omega_{11}\prm = 0 + 0 + 0 + \fk[]\prm \cdot \gamma_{11}\prm \cdot \omega_{11} \\
}

\begin{table} \caption{Patterns for certain or impossible events}
 \renewcommand{\arraystretch}{1.8}
\begin{tabular} {|p{6cm}|p{8cm}|}
\hline
\bf{Certain Events} & \bf{Impossible Events} \\
\hline \hline
$p(\Xk[]\prm = 0 \, \vert \, \Xk[] = 0, \ldots ) = 1$ & $p(\ldots \, \vert \, \Xk[]\prm = 1, \Xk[] = 0, \ldots ) = 0$ \\
$p(\Ck[]\prm = 0 \, \vert \, \ldots, \Ck[] = 0 ) = 1$ & $p(\Xk[]\prm = 1 \, \vert \, \Xk[] = 0, \ldots ) = 0$ \\
 & $p(\Ck[]\prm = 1 \, \vert \, \ldots, \Ck[] = 0 ) = 0$\\
\hline
\end{tabular} \label{tab:certain-or-impossible-events}
\end{table}

\up

Table \ref{tab:certain-or-impossible-events} reports the patterns concerning co-occuring events that are certain or impossible to occur (all probabilities marked as 1 or 0 can be acribed to one of these patterns). They are based on the following tautological implications: 
\ALIGNX{
\Xk[] = 0 \models \Xk[]\prm = 0 \\
\Ck[] = 0 \models \Ck[]\prm = 0
}

Table \ref{tab:approximations} reports the patterns concerning the approximations made while deriving the joint probability $p(\Xk,\Ck)$, the underlying hypothesis being that a high correlation holds between the involved classifiers and the corresponding oracles.
%
%In other words, the more a classifier behaves like an oracle the more the difference between the approximations and the correct values are expected to be negligible.
%
In particular, in presence of co-occurring events such as $\langle \Xk[] = 1, \Ck[] = 1 \rangle$, this assumption permits to disregard $\Xk[] = 1$ or $\Ck[] =1$.

\begin{table} [h!] \caption{Approximation Patterns} \renewcommand{\arraystretch}{1.8}
\begin{tabular} {|p{6cm}|p{8cm}|}
\hline
\bf{Pattern} & \bf{Approximation}\\
\hline \hline
$p(\Ck[]\prm = j \, \vert \, \Xk[]\prm = i, \Xk[] = 1, \Ck[] = 1)$ & $p(\Ck[]\prm = j \, \vert \, \Xk[]\prm = i, \Ck[] = 1) = \gamma_{ij}\prm, \quad i,j = 0,1$ \\
$p(\Xk[]\prm = 1 \, \vert \, \Xk[] = 1, \Ck[] = 1)$ & $p(\Xk[]\prm = 1 \, \vert \, \Xk[] = 1) = \fk[]\prm$  \\
$p(\Xk[]\prm = 0 \, \vert \, \Xk[] = 1, \Ck[] = 1)$ & $p(\Xk[]\prm = 0 \, \vert \, \Xk[] = 1) = \nfk[]\prm$  \\
\hline
$p(\Ck[]\prm = j \, \vert \, \Xk[]\prm = 0, \Xk[] = 0, \Ck[] = 1)$ & $p(\Ck[]\prm = j \, \vert \, \Xk[]\prm = 0, \Ck[] = 1) = \gamma_{0j}\prm, \quad j=0,1$ \\
$p(\Xk[]\prm = 1 \, \vert \, \Xk[] = 1, \Ck[] = 0)$ & $p(\Xk[]\prm = 1 \, \vert \, \Xk[] = 1) = \fk[]\prm$\\
$p(\Xk[]\prm = 0 \, \vert \, \Xk[] = 1, \Ck[] = 0)$ & $p(\Xk[]\prm = 0 \, \vert \, \Xk[] = 1) = \nfk[]\prm$\\
\hline
\end{tabular} \label{tab:approximations}
\end{table}

Other approximations have been made by exploiting also the total probability law. As an example, let us assume that we want to find an approximation for:

\up

\ALIGNX{
p(\Ck[]\prm = 0 \, \vert \, \Xk[]\prm = 0, \Xk[] = 0, \Ck[] = 1)
}

\down

\noindent We know that, by hypothesis:

\up

\ALIGNX{
\gamma_{00}\prm = p(\Ck[]\prm = 0 \, \vert \, \Xk[]\prm = 0, \Ck[] = 1)
}

\noindent Hence, we can write:

\up

\ALIGNX{
\gamma_{00}\prm = p(\Ck[]\prm = 0, \Xk[] = 0 \, \vert \, \Xk[]\prm = 0, \Ck[] = 1) + p(\Ck[]\prm = 0, \Xk[] = 1 \, \vert \, \Xk[]\prm = 0, \Ck[] = 1)
}

\down 

\noindent With $\alpha \mathdef p(\Xk[] = 1 \, \vert \, \Xk[]\prm = 0, \Ck[] = 1)$ and recalling that $\Xk[] = 1$ and $\Ck[] = 1$ are highly correlated by hypothesis, we can write:

\ALIGNX{
\gamma_{00}\prm = p(\Ck[]\prm = 0 \, \vert \, \Xk[]\prm = 0, \Xk[] = 0, \Ck[] = 1) \cdot (1 - \alpha) + \underbrace{p(\Ck[]\prm = 0 \, \vert \, \Xk[]\prm = 0, \Xk[] = 1, \Ck[] = 1)}_{\approx \gamma_{00}\prm} \cdot \alpha
}

\noindent Which yields:

\Up

\ALIGNX{
p(\Ck[]\prm = 0 \vert \Xk[]\prm = 0, \Xk[] = 0, \Ck[] = 1) \approx \gamma_{00}\prm
}

\down

%As a final note, let us point out that,
%although approximations have been done under the assumption that the domains of classifiers nearly coincide with the domains of the corresponding categories,
%preliminary results from experiments performed with a subset of the Reuters taxonomy confirm their robustness also in the event that the domains of classifiers are not highly correlated with the domains of the corresponding categories.

%\noindent \fbox{\rc{This result is not surprising, although it may hyde subtle issues ... NO TIME, NOW}}

% END APPENDIX A.

\newpage

%\section*{Appendix B.} \label{app:homomorphism}

\milestone{APPENDIX B.}

\section*{Appendix B. Deriving the $\bf{\Psi}$ Homomorphism} \label{appendix:homomorphism}

%\section{Deriving the $\bf{\Psi}$ Homomorphism} \label{appendix:homomorphism}

Given a taxonomy $\taxonomy$, it is well known that the closure of $\emh{C}$ under the Kleene star (i.e., $\kleenestar{C}$) yields a monoid, with:

\begin{enumerate}
\item {Closure (wrt the operator ``$+$'')}:\\
$\forall \pi_1, \pi_2 \in \kleenestar{C}: \pi_1 + \pi_2 \in \kleenestar{C}$ \down
\item {Associativity (wrt the operator ``$+$'')}:\\
 $\forall \pi_1, \pi_2, \pi_3 \in \kleenestar{C}: (\pi_1 + \pi_2) + \pi_3 = \pi_1 + (\pi_2+\pi_3)$ \down
\item {Neutral element (empty string $\lambda$)}:\\
 $\forall \pi \in \kleenestar{C}: \pi + \lambda = \lambda + \pi = \pi$ 
\end{enumerate}

In the event that $\Psi$ is a homomorphism, also the set of normalized confusion matrices $\emh{M} \equiv [0,1]^{4}$ is a monoid, as a homomorphism is expected to preserve the structure while mapping $\kleenestar{C}$ to $\emh{M}$.
Let us verify that $\Psi$ is a homomorphism by checking whether the space $\emh{M}$ is a monoid, with $``+'' \rightarrow ``\oplus''$ and $\lambda \rightarrow \neutral$:

\begin{enumerate}
\item {Closure (wrt the operator ``$\oplus$'')}:\\ 
 $\forall a,b \in \emh{M}: a \oplus b \in \emh{M}$ \down 
\item {Associativity (wrt the operator ``$\oplus$'')}:\\
 $\forall a,b,c \in \emh{M}: (a \oplus b) \oplus c = a \oplus (b \oplus c)$ \down 
\item {Neutral element (neutral classifier $\neutral$)}:\\ 
 $\forall a \in \kleenestar{C}: a \oplus \neutral = \neutral \oplus a = a$ 
\end{enumerate}

\noindent \emph{Proof.}

\begin{enumerate}

% CLOSURE
\item {Closure under ``$\oplus$''}:
$\alpha, \beta \in \emh{M} \Rightarrow \alpha \oplus \beta \in \emh{M}$

\ALIGNX {
	\alpha \oplus \beta =
	\begin{bmatrix}
	  \alpha_{00} + \alpha_{01} \cdot \beta_{00} & \alpha_{01} \cdot \beta_{01} \\ 
	  \alpha_{10} + \alpha_{11} \cdot \beta_{10} & \alpha_{11} \cdot \beta_{11}
	\end{bmatrix}
}

\noindent where
\up
\ALIGNX {
	0 \le {\left( \alpha \oplus \beta \right)}_{00} \le \alpha_{00} + \alpha_{01} = 1, \quad \quad &
	0 \le {\left( \alpha \oplus \beta \right)}_{01} \le \alpha_{01} \le 1 \\
	0 \le {\left( \alpha \oplus \beta \right)}_{10} \le \alpha_{10} + \alpha_{11} = 1, \quad \quad &
	0 \le {\left( \alpha \oplus \beta \right)}_{11} \le \alpha_{11} \le 1
}

\noindent Moreover:
\ALIGNX{
{\left( \alpha \oplus \beta \right)}_{00} + {\left( \alpha \oplus \beta \right)}_{01}
&= \left( \alpha_{00} + \alpha_{01} \cdot \beta_{00} \right) + \alpha_{01} \cdot \beta_{01}
  = \alpha_{00} + \alpha_{01} \cdot \left( \beta_{00} + \beta_{01} \right) = 1 \\
{\left( \alpha \oplus \beta \right)}_{10} + {\left( \alpha \oplus \beta \right)}_{11}
&= \left( \alpha_{10} + \alpha_{11} \cdot \beta_{10} \right) + \alpha_{11} \cdot \beta_{11}
  = \alpha_{10} + \alpha_{11} \cdot \left( \beta_{10} + \beta_{11} \right) = 1
}

% ASSOCIATIVITY
\Down
\item {Associativity under ``$\oplus$''}:
$\left( \alpha \oplus \beta \right) \oplus \gamma = \alpha \oplus \left( \beta \oplus \gamma \right)$

\ALIGNX {
	\left( \alpha \oplus \beta \right) \oplus \gamma
	&= \OOOplus{\alpha}{\beta} \oplus \cmx{\gamma}\\
	&=
	\begin{bmatrix}
	  \alpha_{00} + \alpha_{01} \cdot \beta_{00} & 0 \\ 
	  \alpha_{10} + \alpha_{11} \cdot \beta_{10} & 0
	\end{bmatrix}
	+
	\begin{bmatrix}
	  \alpha_{01} \cdot \beta_{01} & 0 \\ 
	  0 & \alpha_{11} \cdot \beta_{11}
	\end{bmatrix}
	\cdot \cmx { \gamma } \\
         &=
	\begin{bmatrix}
         \alpha_{00} + \alpha_{01} \cdot \beta_{00} + \alpha_{01} \cdot \beta_{01}  \cdot \gamma_{00} &
         \alpha_{01} \cdot \beta_{01} \cdot \gamma_{01} \\
         \alpha_{10} + \alpha_{11} \cdot \beta_{10} + \alpha_{11} \cdot \beta_{11}  \cdot \gamma_{10} &
         \alpha_{11} \cdot \beta_{11}  \cdot \gamma_{11}
	\end{bmatrix} \\
%}
%
%\up
%
%\ALIGNX {
	\alpha \oplus \left( \beta \oplus \gamma \right)
	&=
	\cmx { \alpha }
	\oplus
	\begin{bmatrix}
	  \beta_{00} + \beta_{01} \cdot \gamma_{00} & \beta_{01} \cdot \gamma_{01} \\ 
	  \beta_{10} + \beta_{11} \cdot \gamma_{10} & \beta_{11} \cdot \gamma_{11}
	\end{bmatrix} \\
	& =
	\outercmx {\alpha_{00}}{\alpha_{10}}
	+
	\diagcmx { \alpha_{01} } { \alpha_{11} }
	\cdot
	\begin{bmatrix}
	  \beta_{00} + \beta_{01} \cdot \gamma_{00} & \beta_{01} \cdot \gamma_{01} \\ 
	  \beta_{10} + \beta_{11} \cdot \gamma_{10} & \beta_{11} \cdot \gamma_{11}
	\end{bmatrix}\\
         &=
	\begin{bmatrix}
         \alpha_{00} + \alpha_{01} \cdot \beta_{00} + \alpha_{01} \cdot \beta_{01}  \cdot \gamma_{00} & &
         \alpha_{01} \cdot \beta_{01} \cdot \gamma_{01} \\
         \alpha_{10} + \alpha_{11} \cdot \beta_{10} + \alpha_{11} \cdot \beta_{11}  \cdot \gamma_{10} & &
         \alpha_{11} \cdot \beta_{11}  \cdot \gamma_{11}
	\end{bmatrix}
}

\item{Neutral element $\neutral$}:
$\alpha \in \emh{M} \Rightarrow \alpha \oplus \neutral = \neutral \oplus \alpha \equiv \alpha, \;$ with $\; \; \neutral = \neutralcmx$

The neutral element $\neutral$ corresponds to a classifier that accepts and passes down all its input data (i.e., FP and TP).
It is easy to verify that this property holds for the choice made about $\neutral$:

\ALIGNX {
	\alpha \oplus \neutral & =
	\cmx { \alpha } \oplus \neutralcmx =
	\begin{bmatrix} \alpha_{00} & 0 \\ \alpha_{10} & 0 \end{bmatrix}
	+
	\begin{bmatrix} \alpha_{01} & 0 \\ 0 & \alpha_{11} \end{bmatrix}
	\cdot
	\begin{bmatrix} 0 & 1 \\ 0 & 1 \end{bmatrix}\equiv \alpha \\ \\
	\neutral \oplus \alpha & = \neutralcmx \oplus \cmx { \alpha }
	=
	\zerocmx + \identitycmx \cdot \cmx { \alpha } \equiv \alpha
}

\end{enumerate}

% END APPENDIX B.

\end{document}